\pdfoutput=1
\documentclass[journal]{IEEEtran}
\usepackage{cite}
\usepackage{amsmath,amssymb,amsfonts}
\usepackage{algorithmic}
\usepackage{graphicx}
\usepackage{textcomp}
\usepackage{xcolor}
\usepackage{multirow}
\usepackage{makecell}
\usepackage{bm}
\usepackage[caption=false,font=footnotesize]{subfig}
\usepackage{hyperref}

\newcommand{\sectionref}[1]{Section \ref{#1}}

\newcommand{\reftable}[1]{Table~\ref{#1}}
\newcommand{\reffig}[1]{Fig.~\ref{#1}}
\newcommand{\reffigs}[2]{Figs.~\ref{#1} and \ref{#2}}
\newcommand{\refConf}[1]{Conf.~\ref{#1}}

\newcommand{\refConfs}[2]{Confs.~\ref{#1}--\ref{#2}}

\begin{document}

\title{Promptable Concept Segmentation from Above: Evaluating Zero-Shot Textual and Visual Prompting of SAM~3 in Remote Sensing}

\author{
Mohammad Dabaja, Turgay Celik
\thanks{Both authors are affiliated with the Department of ICT,
University of Agder, Grimstad, Norway (e-mail: mohammadkd@uia.no; turgay.celik@uia.no).}
\thanks{Corresponding author: Mohammad Dabaja (e-mail: mohammad.dabaja.edu@gmail.com).}
}

\maketitle

\begin{abstract}
Large-scale foundation models such as the Segment Anything Model~3 (SAM~3) promise a transition toward open-vocabulary, training-free computer vision. However, their capacity to generalize out-of-distribution to the top-down geometric structures of Earth Observation imagery remains largely unquantified. We present a multi-task empirical evaluation of SAM~3 across remote sensing scene classification, object detection, and instance segmentation, without any fine-tuning. For scene classification, we repurpose its decoupled binary presence head as a standalone zero-shot classifier. For detection and segmentation, we compare textual and visual prompts, where a visual prompt reveals one true instance of the target class in the image being segmented. Text alone performs poorly, while adding a visual prompt raises accuracy sharply, and combining both raises or lowers accuracy depending on the task. We also compare SAM~3's text-only performance against several general-domain, open-vocabulary detection and segmentation models under the same evaluation protocol, and find that it leads all of them on both benchmarks. Its accuracy without a visual prompt nonetheless remains limited, motivating parameter-efficient fine-tuning as a direction for future work.
\end{abstract}


\section{Introduction}
\label{section:Introduction}
\IEEEPARstart{T}{he} emergence of massive vision-language foundation models has transitioned the computer vision field away from dataset-specific, closed-set architectures toward open-vocabulary architectures. Central to this evolution is the concept of promptable segmentation, introduced by the Segment Anything Model (SAM)~\cite{kirillov2023segment}. SAM shifted from class-specific segmentation to class-agnostic segmentation guided by spatial prompts, a capability subsequently extended to temporal video tracking by SAM~2~\cite{ravi2024sam2}. The deployment of SAM~3~\cite{carion2026sam3} advances this class-agnostic foundation toward a unified task defined as Promptable Concept Segmentation (PCS). By integrating a multimodal decoder and a decoupled presence head, SAM~3 separates global semantic recognition from localized spatial bounding. This decoupled architecture unifies zero-shot conceptual reasoning with precise mask generation. However, initial benchmark data reveal a critical limitation. While SAM~3 achieves 56.4 AP on general-domain datasets like COCO, its zero-shot performance drops to 15.2 AP on the RF-100VL benchmark~\cite{carion2026sam3}, which comprises 100 specialized visual domains. This performance gap indicates that while SAM~3 acts as an effective generalist, its direct application to highly specialized fields, such as Earth Observation (EO) and Remote Sensing (RS), requires objective evaluation to determine if costly domain-specific fine-tuning is necessary.

The direct application of general-domain foundation models to remote sensing imagery is obstructed by structural, geometric, and semantic domain gaps. Unlike standard consumer photography, satellite and aerial imagery are characterized by a nadir (top-down) perspective, multi-scale distributions, dense background clutter, and an absence of canonical orientation~\cite{liu2023remoteclip}. This spatial anisotropy induces a cross-modal alignment failure within the model's multimodal fusion layers. Because standard vision-language models are optimized contrastively over ground-level image-caption pairs, their text encoders map concepts to latent vectors based on horizontal viewpoints. For example, the linguistic embedding of a ``school'' is implicitly tied to its vertical facades and windows. However, when viewed from an overhead satellite perspective, these objects collapse into flat, two-dimensional geometric patches. Consequently, the multimodal decoder fails to reconcile ground-level textual expectations with top-down visual features. This creates semantic blind spots, despite the model possessing a robust spatial feature extraction backbone.

To address this cross-modal alignment deficit, existing remote sensing frameworks typically rely on supervised fine-tuning over labeled aerial imagery. While this calibration step can encode dataset-specific contextual and spatial priors, it also requires substantial labeled data collection for each target domain and constrains the resulting model to a fixed, closed vocabulary of trained categories. This raises a more fundamental question about whether a general-domain foundation model can bridge this cross-modal alignment gap without any fine-tuning at all.

To answer this question and establish a performance baseline, this study evaluates SAM~3 under zero fine-tuning. We analyze the multimodal processing of the model across the AID~\cite{xia2017aid}, DIOR~\cite{li2020dior}, and iSAID~\cite{zamir2019isaid} datasets by varying the input prompts. These benchmarks serve as the established standards for evaluation in the remote sensing literature. We adapt SAM~3's decoupled architecture by utilizing its binary presence head as a standalone global scene classifier. Furthermore, we define a matrix of five prompt configurations (\refConfs{config:text}{config:textbox_filt}) to map how SAM~3's multimodal decoder processes remote sensing input prompts. These configurations isolate textual prompts, visual prompts, multimodal prompt combinations, and oracle negative filtering. Because to our knowledge, no prior work reports a general-domain, open-vocabulary detection or segmentation model evaluated directly on DIOR or iSAID under standard zero-shot mAP, we additionally benchmark SAM~3 against four such model families (OWLv2~\cite{minderer2023owlv2}, Grounding DINO~\cite{liu2023grounding}, YOLO-World~\cite{yoloworld2024}, and YOLOE~\cite{wang2025yoloe}), establishing the first direct, same-protocol comparison of this kind on these two datasets.

The primary contributions of this work are summarized as follows:
\begin{itemize}
    \item We introduce a structural adaptation of SAM~3, utilising its decoupled binary presence head as a standalone zero-shot scene classifier to evaluate the global semantic awareness of its multimodal decoder.
    \item We conduct a prompt ablation study using five configurations across object detection and instance segmentation. This examines how the visual and textual prompts interact within the multimodal decoder, showing that visual prompts achieve strong spatial alignment while the effect of adding text to them depends on the task.
    \item We benchmark SAM~3 against four general-domain, open-vocabulary detection and segmentation model families (OWLv2, Grounding DINO, YOLO-World, and YOLOE) directly on the DIOR and iSAID datasets using standard zero-shot mAP.
    \item We provide a class-by-class diagnostic analysis detailing how target size, visual complexity, and geometric invariance influence the performance of zero-shot geospatial foundation models. This analysis identifies a resolution bottleneck regarding small-scale feature representation.
\end{itemize}

The remainder of this paper is organized as follows. \sectionref{section:Related Work} reviews the literature on vision-language models, SAM promptable segmentation, and general-domain open-vocabulary detection and segmentation models. \sectionref{section:Methodology} describes the benchmark datasets, the proposed adaptation of SAM~3 for zero-shot scene classification, the prompting configurations, and the evaluation protocol. \sectionref{section:Experimental Results and Discussion} presents the experimental results on scene classification, object detection, and instance segmentation, together with quantitative and qualitative analyses. Finally, \sectionref{section:Conclusion and Future Work} summarizes the main findings and outlines directions for future research.

\section{Related Work}
\label{section:Related Work}

\subsection{Vision-Language Models and Zero-Shot Scene Classification}
The development of Vision-Language Models (VLMs), such as OpenAI CLIP~\cite{radford2021clip}, established the foundation for open-vocabulary scene understanding through large-scale contrastive learning. By optimizing a bi-directional InfoNCE loss, these models map visual and textual feature vectors into a shared latent embedding space. While standard CLIP demonstrates robust scene comprehension on ground-level photography, its direct application to Earth Observation (EO) imagery exposes a semantic domain gap~\cite{liu2023remoteclip}. Satellite imagery is characterized by a nadir (top-down) perspective, dense multi-scale object distributions, and high inter-class similarity. For example, distinguishing between ``medium residential'' and ``dense residential'' zones requires fine-grained feature extraction. Consequently, the visual feature vectors extracted from overhead perspectives are shifted in the latent space, failing to align with their corresponding textual feature vectors~\cite{georsclip2024}.

To address this perspective mismatch, recent models employ domain-specific continuous pre-training. Architectures such as RemoteCLIP~\cite{liu2023remoteclip}, GeoRSCLIP~\cite{georsclip2024}, and RSDiX~\cite{rsdixclip2025} adapt foundation models by fine-tuning on large datasets using rule-based annotations or self-distillation techniques. While this targeted contrastive alignment improves whole-scene aerial classification accuracy, these VLMs are constrained by their global image-level objectives. Consequently, they lack the dense prediction capabilities required for bounding box localization and precise mask generation.

\subsection{The Evolution of Promptable Architectures: From SAM to SAM~3}

\begin{figure*}[h]
    \centering
    \includegraphics[width=\textwidth]{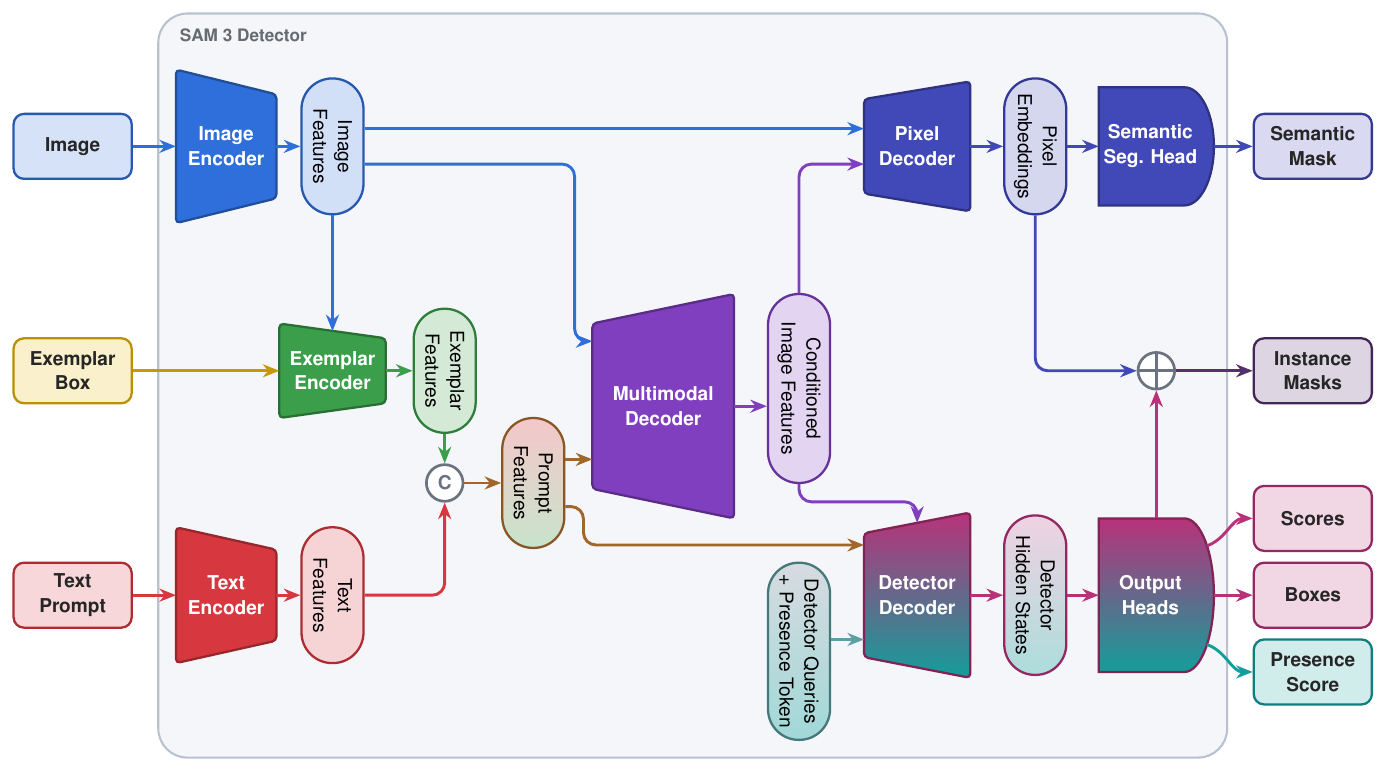} 
    \caption{Architecture of the SAM~3 image-level detector, adapted from~\cite{carion2026sam3}. The exemplar encoder combines the box prompt with the image features, and the resulting exemplar features are concatenated with the text features to form the prompt features. The Multimodal Decoder conditions the image features on these prompt features. The conditioned features then feed a Pixel Decoder for the semantic and instance masks, and a Detector Decoder whose learned queries and presence token produce the scores, boxes and global presence score. Colours follow the information flow, with blue, red and yellow marking the image, text and exemplar inputs and blended colours marking where these streams are combined. The video tracker is omitted since this work considers images only.}
    \label{fig:sam3_architecture}
\end{figure*}

To decouple spatial localization from closed-set classification, the original Segment Anything Model (SAM)~\cite{kirillov2023segment} introduced promptable visual segmentation. It utilized a Vision Transformer (ViT) image encoder paired with a prompt-driven mask decoder. However, its core design was class-agnostic. While it achieved zero-shot localization guided by spatial prompts (points or boxes), it lacked a semantic latent space to comprehend and segment based on open-vocabulary textual concepts. Its successor, SAM~2~\cite{ravi2024sam2}, extended this promptable architecture into the temporal domain to enable continuous video object tracking. Yet, despite its temporal advancements, it operated primarily as a class-agnostic localizer lacking the explicit multimodal alignment required for zero-shot semantic recognition.

SAM~3~\cite{carion2026sam3} introduces an architecture designed to address Promptable Concept Segmentation (PCS). As illustrated in \reffig{fig:sam3_architecture}, SAM~3 transitions from a purely spatial system to a unified multimodal framework. It expands the promptable interface by introducing a dedicated Text Encoder and an Exemplar Encoder to process text and visual prompts, respectively. These extracted prompt features, alongside the base Image Features, are fed into the Multimodal Decoder to generate Conditioned Image Features. The architecture then bifurcates. A Pixel Decoder upsamples the Conditioned Image Features to generate high-resolution per-pixel embeddings. These embeddings are routed to a Semantic Segmentation Head to produce a broad Semantic Mask. Simultaneously, a Detector Decoder processes the Conditioned Image Features, the original prompt features, and a fixed set of learned Detector Queries paired with a Presence Token. This design is heavily inspired by the DEtection TRansformer (DETR) paradigm~\cite{carion2020end}, where a fixed number of learned object queries interact with encoded features via cross-attention. Through these mechanisms, the Output Heads translate the decoded states into localized bounding boxes and final semantic scores. Crucially, to generate precise Instance Masks, the pixel-level embeddings from the Pixel Decoder are mathematically fused with the object-level queries from the Output Heads.

A fundamental challenge in open-vocabulary segmentation is the architectural trade-off between semantic recognition and spatial localization. Aggregating global context to resolve semantic ambiguity often requires feature pooling, which degrades the fine-grained spatial details required for precise mask boundaries. 

To resolve this architectural trade-off, SAM~3 introduces a decoupled Presence Head. Instead of requiring the local detector queries to independently predict both the global existence of a concept and its exact boundaries, the architecture decomposes the prediction. A shared, learned Presence Token independently evaluates the Global Presence Score, which represents the probability that the requested concept exists anywhere in the scene. The final match probability for each individual query is factored as:
\begin{equation}
\begin{aligned}
p(\text{query}_i \text{ matches NP}) &= p(\text{query}_i \text{ matches NP} \mid \text{NP appears}) \\
&\quad \cdot p(\text{NP appears})
\end{aligned}
\label{eq:sam3_prob}
\end{equation}
Conditioned on this global presence, the local object queries can focus strictly on foreground-background localization without being constrained by global contextual demands. This decoupled approach prevents localized queries from being overwhelmed by global classification tasks, thereby improving detection precision and reducing false positives in dense environments.

\subsection{General-Domain Open-Vocabulary Detection and Segmentation Models}
\label{subsec:related_ovd}
Open-vocabulary detection has developed rapidly since 2023, with each new model contributing a distinct mechanism for connecting language and visual localization. Grounding DINO~\cite{liu2023grounding} combines a DETR-style transformer detector with grounded language pre-training, encoding text prompts at the sub-sentence level to prevent semantic interference when multiple category names are queried together. Released shortly after, OWLv2~\cite{minderer2023owlv2} takes a different route to the same goal. Rather than changing the detector architecture, it introduces OWL-ST, a self-training recipe that generates pseudo-box annotations from web-scale image-text pairs, scaling training data far beyond what human box annotation allows. In 2024, YOLO-World~\cite{yoloworld2024} brought open-vocabulary detection into a real-time setting, building on the single-stage YOLO architecture with a re-parameterizable vision-language path aggregation network and a prompt-then-detect strategy, trading some accuracy for substantially faster inference. YOLOE~\cite{wang2025yoloe} followed in 2025, unifying detection and instance segmentation within a single model and producing masks natively rather than relying on a separate segmentation step. Most recently, SAM~3~\cite{carion2026sam3} was released in 2026, separating concept recognition from spatial localization through an explicit presence head, the mechanism this study builds on throughout.

Because Grounding DINO, OWLv2, and YOLO-World output bounding boxes only, a common practice for evaluating them as segmenters is to pass their predicted boxes into SAM~\cite{kirillov2023segment} as box prompts, converting detections into instance masks through a two-stage detect-then-segment pipeline. YOLOE and SAM~3 differ from this pattern by producing masks directly, in a single forward pass.

\section{Methodology}
\label{section:Methodology}

\subsection{Datasets}
To evaluate SAM~3's zero-shot capabilities across diverse remote sensing tasks, we selected three distinct benchmark datasets, covering classification, detection, and segmentation:

\begin{itemize}
    \item The \textbf{Aerial Image Dataset (AID)}~\cite{xia2017aid} is a benchmark for aerial scene classification, comprising 10,000 multi-source Google Earth images distributed across 30 semantic scene categories, including airport, farmland, industrial, and residential areas. Each image has a fixed spatial dimension of $600 \times 600$ pixels, with spatial resolution ranging from 0.5 m to 8 m per pixel depending on the source imagery. The dataset is partitioned into training and test subsets at a standard 50\%/50\% ratio. Key challenges include substantial intra-class variation and inter-class similarity arising from the multi-source nature of the imagery.
 
    \item The \textbf{Object Detection in Optical Remote Sensing Images Dataset (DIOR)}~\cite{li2020dior} is a benchmark for object detection in optical remote sensing imagery, comprising 23,463 images of $800 \times 800$ pixels with 192,472 annotated instances across 20 object categories labeled with horizontal bounding boxes. Spatial resolution ranges from 0.5 m to 30 m per pixel, with images sourced from over 80 countries. The dataset is partitioned into training, validation, and test subsets of 5,862, 5,863, and 11,738 images, respectively. Key challenges include wide object-scale variation, high inter-class similarity, and significant diversity in imaging conditions across weather, seasons, and viewpoints.

    \item The \textbf{Instance Segmentation in Aerial Images Dataset (iSAID)}~\cite{zamir2019isaid} is a benchmark for instance segmentation in aerial imagery, comprising 2,806 high-resolution images with 655,451 densely annotated object instances across 15 categories, combining instance-level detection with pixel-level segmentation masks. Image sizes range from $800 \times 800$ to $4,000 \times 13,000$ pixels, collected from multiple sensors and platforms at varying spatial resolutions. The dataset is partitioned into training, validation, and test subsets of 1,411, 458, and 937 images, respectively. Key challenges include large object-scale variation, high instance density per image, and the prevalence of small objects.
\end{itemize}

For iSAID, high-resolution images from the validation split (458 images) were cropped into uniform sub-patches using a sliding window, since iSAID's test split has no publicly available annotations. Each image was partitioned into $800 \times 800$ pixel patches with a stride of 600 pixels, giving a 200-pixel (25\%) overlap between adjacent patches. Boundary regions were handled by shifting the final window flush with the image border rather than zero-padding. This produced 9,512 patches, of which 6,050 contain at least one annotated instance. All 9,512 patches, including those with no annotated instance, are used for evaluation across every model and configuration in this study. DIOR images, by contrast, are used at their native resolution, with no resizing or tiling, since no equivalent preprocessing was required.

Qualitative examples throughout this study were selected by hand to illustrate patterns identified in the quantitative results, prioritizing images that exhibit multiple co-occurring behaviors, varied object classes, and a sufficient number of instances for visual interpretation.

\subsection{Adapting SAM~3 for Zero-Shot Scene Classification}
\label{meth:cls}
Because SAM~3 is a promptable binary segmentation model designed for localized mask generation, it requires structural adaptation for whole-image scene classification. We utilise its decoupled architecture, specifically the independent Presence Head, to evaluate classification in a zero-shot setting. 

Let $\mathbf{I}$ represent an input image and $\mathcal{C} = \{c_1, c_2, \dots, c_{N}\}$ denote the set of $N$ target concepts in a given dataset (where $N = 30$ for the AID dataset). For each concept $c \in \mathcal{C}$, we independently query SAM~3 to yield two scalar outputs:
\begin{enumerate}
    \item \textbf{Presence Score ($P_c$):} The global token probability indicating the existence of concept $c$ within $\mathbf{I}$.
    \item \textbf{Conditional Localisation Score ($L_c$):} The highest localisation score assigned by the Output Heads, conditioned on the presence of the concept.
\end{enumerate}

To establish the final predicted image class $\hat{y}$, we formally evaluate three decision functions. The baseline functions rely on standalone probabilities: 

\begin{equation}
\hat{y}_{\text{pres}} = \arg\max_{c \in \mathcal{C}} P_c
\label{eq:decision_pres}
\end{equation}
\begin{equation}
\hat{y}_{\text{local}} = \arg\max_{c \in \mathcal{C}} L_c
\label{eq:decision_logit}
\end{equation}

To fuse global semantic context with localized geometric confidence, we compute a combined Confidence Score ($S_c$). This formulation natively reconstructs SAM~3's decoupled query match probability \eqref{eq:sam3_prob} by multiplying the global presence probability by the local conditional confidence:

\begin{equation}
S_c = P_c \cdot L_c
\label{eq:confidence_score}
\end{equation}

This yields the final decision function: 
\begin{equation}
\hat{y}_{\text{combined}} = \arg\max_{c \in \mathcal{C}} S_c
\label{eq:combined_classification}
\end{equation}

Besides Top-1 accuracy, we report Top-$k$ accuracy, where an image is counted as correct when its ground-truth class is among the $k$ concepts with the highest score under the respective decision function.

\subsection{Formalizing Prompt Modalities and Negative Filtering}
\label{subsec:prompt_modalities}
For open-vocabulary object detection and instance segmentation, we evaluate the model by prompting each target class independently per image. Following SAM~3, we call the input that specifies the target a concept prompt, which can be textual, visual, or both. To systematically isolate the impact of different prompt modalities, we define the input prompt set $\mathcal{Q}_c$ for a target class $c$. The textual prompt $t_c$ is a short noun phrase taken from the class name. The visual prompt $b_c$ is a bounding box with a positive label placed on one instance of the class, which SAM~3 refers to as an image exemplar. Unlike the single-instance visual prompts of SAM~1 and SAM~2, SAM~3 uses it to find all instances of the concept in the image. When a class is present in an image, $b_c$ is taken from one of its ground-truth instances in that same image, chosen at random. This follows the SAM~3 paper, which also samples image exemplars at random from the ground truth. No separate support image is used, since SAM~3 takes its exemplars on the image being segmented. Because a single random draw of $b_c$ could by chance land on an easy or a hard instance, every configuration that uses the visual prompt (Box-Only, Text + Box, and Text + Box + Oracle) was run independently seven times, each with a different random seed governing the exemplar draw for every class and image, on both DIOR and iSAID. We report the mean and sample standard deviation across these seven runs for all detection and segmentation metrics in this study. Text-Only and Text + Oracle use no visual prompt and are therefore deterministic given the fixed model weights, so each was run once. Because the visual prompt reveals the location and appearance of one true instance, configurations that use it measure how well the model localizes and outlines objects when told what to look for, and we do not compare them with the text-only baselines of other models. All configurations are zero-shot in that no weights are fine-tuned or updated on remote sensing data, and they differ only in which prompts the model receives. We establish three core experimental modalities:

\begin{enumerate}
    \item \label{config:text} \textbf{Text-Only (Conf.~1):} Zero-shot semantic inference for every class in every image, where $\mathcal{Q}_c = \{t_c\}$.
    \item \label{config:box} \textbf{Box-Only (Conf.~2):} Visual inference bypassing the text encoder, where $\mathcal{Q}_c = \{b_c\}$. A box exists only for classes present in the image, so only those classes are queried.
    \item \label{config:textbox} \textbf{Text + Box (Conf.~3):} Multimodal inference combining both prompt types, where $\mathcal{Q}_c = \{t_c, b_c\}$ for classes present in the image and $\mathcal{Q}_c = \{t_c\}$ for absent classes.
\end{enumerate}

To evaluate the limit of the decoupled presence head and track the impact of false-positive noise, we introduce an oracle ground-truth negative filtering. Let ${f}_c \in \{0, 1\}$ be a binary indicator function that returns $1$ if at least one instance of concept $c$ is present in the ground truth of image $\mathbf{I}$, and $0$ otherwise. The oracle Confidence Score $\tilde{S}_c$ is computed as:
\begin{equation}
\tilde{S}_c = S_c \cdot {f}_c
\label{eq:oracle_filter}
\end{equation}
This formulation suppresses the model's output by forcing the confidence score to absolute zero for any category not represented in the image background. This negative suppression mechanism yields two additional evaluation configurations:
\begin{enumerate}
    \setcounter{enumi}{3} 
    
    \item \label{config:text_filt} \textbf{Text + Oracle filtered (Conf.~4):} \refConf{config:text} subjected to oracle negative suppression.
    \item \label{config:textbox_filt} \textbf{Text + Box + Oracle filtered (Conf.~5):} \refConf{config:textbox} subjected to oracle negative suppression.
\end{enumerate}
Comparing the filtered configurations' output against the unfiltered baselines quantifies the model's susceptibility to false-positive when prompted with absent concepts. Because the oracle uses ground truth, Conf.~4 and 5 are upper bounds on what a perfect presence predictor could add and do not represent practical inference. Box-Only shares this property, since it can only be run on classes present in the image.

\subsection{Evaluation Protocol for Open-Vocabulary Object Detection and Instance Segmentation}
\label{subsec:eval_protocol}
Performance for both object detection and instance segmentation is quantified using Average Precision (AP), following the standard COCO definition: precision averaged across recall levels and across Intersection over Union (IoU) thresholds from 0.50 to 0.95, denoted AP$_{50:95}$, or at the single 0.50 IoU threshold, denoted AP$_{50}$. AP$_S$, AP$_M$, and AP$_L$ denote the same quantity restricted to small, medium, and large ground-truth instances by pixel area, following the COCO size definition. Throughout this paper, AP denotes a single class's score, and mAP denotes the mean of AP across classes, so mAP$_{50:95}$, mAP$_{50}$, mAP$_S$, mAP$_M$, and mAP$_L$ denote the class-averaged values reported for each configuration as a whole. We confirmed that every reported mAP value is equivalent to the simple average of the per-class AP values shown in the corresponding class-wise table, since every class has ground-truth instances present in the evaluated set.

The cross-modal comparison asks one question, namely what changes when a textual prompt is added to a box prompt. Because Box-Only and Text + Box + Oracle score the same present classes and share the same exemplar in every run, their difference isolates the contribution of the text, and we report it as the mean over the seven exemplar seeds of Text + Box + Oracle minus the mean of Box-Only. A single difference of two means does not show how reliable it is or where it comes from, so we add two views of it. The image-wise view is a paired bootstrap that resamples test images with replacement 1000 times, applies the same resample to both configurations and to all seven seeds, recomputes AP from the resampled per-image evaluation results, and takes the 2.5th and 97.5th percentiles of the change as a 95\% interval. Pairing removes the variation that comes from some images being harder than others, so the interval reflects only the effect of the text and tells us whether it is distinguishable from image-sampling noise. The class-wise view applies the same procedure to each class separately, and shows whether the change is spread evenly across classes or concentrated in a few of them.

SAM~3 is evaluated using its single publicly released image checkpoint (\texttt{facebook/sam3}). As with OWLv2, YOLO-World, and YOLOE, no confidence threshold or additional non-maximum suppression is applied to its predictions. Every detection it returns is kept and scored over the full ranked list.

We compare SAM~3 against OWLv2 and Grounding DINO, the same two models used as baselines when evaluating SAM~3 itself in its own paper, and add YOLO-World and YOLOE to bring in more recent open-vocabulary architectures. All four are evaluated on DIOR and iSAID using the same class-name prompting convention as SAM~3's own text prompt $t_c$. For each model we aimed to get the strongest result it could reasonably produce on these two datasets, which took some experimentation, most notably around each model's confidence threshold.

OWLv2, in both its base and ensemble variants, YOLO-World at its small and large sizes, and YOLOE at its small and large sizes were all run with a confidence threshold near zero, since each returns a clean, directly rankable score per detection and Average Precision is computed over the full ranked list rather than a single cutoff. Grounding DINO, in its tiny and base variants, needed a different approach. Its free-text label decoding produced many confidently mislabeled boxes at low thresholds, so we searched a small set of threshold pairs on held-out images and kept the setting that gave the strongest result, a box and text confidence of 0.25 each. Per-category non-maximum suppression at an IoU of 0.5 was applied on top of each model's own internal suppression, and up to 100 detections were kept per image for each category, matching how the COCO evaluation API applies its own detection limit internally.

OWLv2, Grounding DINO, and YOLO-World only produce bounding boxes. Following the detect-then-segment approach described in Section~\ref{subsec:related_ovd}, their predicted boxes were passed to SAM~1 as box prompts, taking a single mask per box with no further post-processing, the same protocol the SAM~3 paper uses when converting baseline detections into masks. YOLOE was evaluated twice on iSAID, once through this same detect-then-SAM pipeline and once using its own segmentation head, which produces a mask directly from the same forward pass as detection, making it the only model besides SAM~3 able to segment in a single step.

\section{Experimental Results and Discussion}
\label{section:Experimental Results and Discussion}

\subsection{Aerial Scene Classification (AID)}
\begin{figure}
    \centering
    \includegraphics[width=\columnwidth]{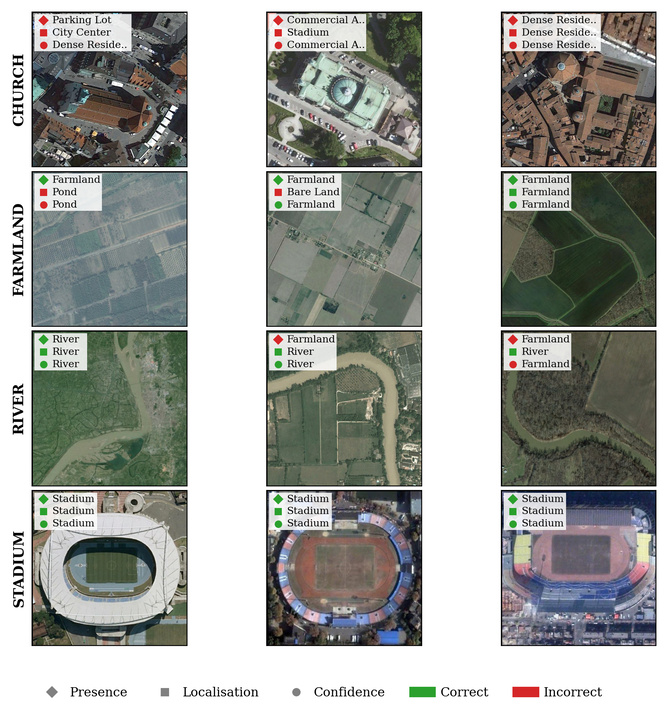} 
    \caption{Qualitative zero-shot scene classification results on selected categories from the AID dataset. Each row displays three distinct image examples for a target class, illustrating different combinations of successful and unsuccessful predictions. Marker shapes correspond to the three evaluated decision functions: the Presence Score ($\hat{y}_{\text{pres}}$), indicated by a diamond ($\diamond$); the Conditional Localisation Score ($\hat{y}_{\text{local}}$), indicated by a square ($\square$); and the Confidence Score ($\hat{y}_{\text{combined}}$), indicated by a circle ($\circ$). Green markers denote correct classifications, while red markers denote incorrect classifications. The rows demonstrate four distinct outcomes: (1) \textit{Church}, where all three decision functions produce incorrect classifications across the provided examples; (2) \textit{Farmland}, where only the decision function based on the Presence Score produces correct classifications; (3) \textit{River}, where only the decision function based on the Conditional Localisation Score produces correct classifications; and (4) \textit{Stadium}, where all three decision functions produce correct classifications across all examples.}
    \label{fig:aid_qualitative}
\end{figure}

We evaluate SAM~3 on the 30-class AID dataset in a zero-shot setup as explained in \ref{meth:cls}.  The macro performance results are consolidated alongside a cross-model baseline evaluation in \reftable{tab:aid_cross_model_comparison}. The baseline performance metrics for OpenAI CLIP L/14 are reported as reproduced by SkyScript \cite{wang2024skyscript} while the metrics for the B/32 and Resnet-50 by RemoteCLIP~\cite{liu2023remoteclip}. The complete, 30-class performance target matrix is provided in \reftable{tab:appendix_aid_exhaustive}. Corresponding qualitative results are illustrated in \reffig{fig:aid_qualitative}.

\begin{table}
\caption{Cross-Model Benchmark Comparison on the AID Dataset. We report zero-shot Top-1 scene classification accuracies for SAM~3 and compare them against existing foundational and remote-sensing-adapted vision-language models. Baseline results for all competing methods are taken directly from their respective original publications or reported benchmark studies. SAM~3 is evaluated using three decision strategies derived from its promptable segmentation architecture: Presence Score, Conditional Localisation Score, and their multiplicative Confidence Score. All methods correspond to zero-shot settings on the AID dataset.}
\label{tab:aid_cross_model_comparison}
\centering
\begin{tabular}{|l|l|c|}
\hline
\textbf{Model Family} & \textbf{Specific Variant} & \textbf{Acc. (\%)} \\
\hline

OpenAI CLIP~\cite{radford2021clip} & ResNet-50 & 57.35 \\
OpenAI CLIP~\cite{radford2021clip} & B/32 & 65.65 \\
OpenAI CLIP~\cite{radford2021clip} & L/14 & 69.25 \\

RemoteCLIP~\cite{liu2023remoteclip} & ResNet-50 & 86.55 \\
RemoteCLIP~\cite{liu2023remoteclip} & B/32 & 91.30 \\
RemoteCLIP~\cite{liu2023remoteclip} & L/14 & 87.90 \\

SkyCLIP-50~\cite{wang2024skyscript} & B/32 & 70.90 \\
SkyCLIP-50~\cite{wang2024skyscript} & L/14 & 71.70 \\

RS-CLIP~\cite{li2023rsclip} & B/32 & 65.48 \\

GeoRSCLIP~\cite{georsclip2024} & B/32 & 73.72 \\
GeoRSCLIP~\cite{georsclip2024} & H/14 & 76.33 \\

RSDiX-CLIP~\cite{rsdixclip2025} & B/32 & \textbf{95.10} \\
RSDiX-CLIP~\cite{rsdixclip2025} & B/16 & 92.60 \\
RSDiX-CLIP~\cite{rsdixclip2025} & L/14 & 91.10 \\

HQRS-CLIP~\cite{he2025hqrsclip} & - & 73.86 \\
\hline

\textbf{SAM~3} & Presence Score \eqref{eq:decision_pres}  & 37.26 \\
\textbf{SAM~3} & Conditional Localisation Score \eqref{eq:decision_logit} & 28.25 \\
\textbf{SAM~3} & Confidence Score \eqref{eq:confidence_score} & \textbf{38.70} \\
\hline
\end{tabular}
\end{table}

\subsubsection{Scoring Head Dynamics (Presence vs. Localisation)} 

As shown in \reftable{tab:aid_cross_model_comparison}, the classification strategy guided by the standalone Presence Score \eqref{eq:decision_pres} achieves an accuracy of 37.26\%, outperforming the Conditional Localisation Score \eqref{eq:decision_logit}, which yields 28.25\%. This performance gap highlights the structural decoupling of the model. The independent presence head evaluates the global existence of the target concept across the entire scene embedding, whereas the output heads verify the presence of the specific conceptual prompt within localized sub-regions of the shared space, conditioned on the concept existing in the image. 

Fusing these outputs via the combined Confidence Score \eqref{eq:combined_classification} yields the highest classification accuracy of 38.70\%, which indicates that these decoupled predictions provide complementary representations. The Localisation score provides conditional localized verification of the concept within the shared embedding space, while the global presence score provides a macro-level prior for the overall scene context. Consequently, this joint decision rule enforces a joint probability constraint that requires a concept to satisfy both global and localized verification criteria simultaneously, improving scene classification performance compared to either standalone evaluation score.

This ordering does not depend on how strictly a prediction is scored. \reftable{tab:aid_topk} extends the evaluation from Top-1 to Top-3, Top-5, and Top-10 accuracy. The Presence Score and the Confidence Score differ by at most 1.44 points at any $k$ (both reach 66.44\% at Top-5), and both stay between 8.8 and 11.7 points above the Conditional Localisation Score throughout. For the two stronger decision functions, accuracy rises from below 40\% at Top-1 to roughly 80\% at Top-10, so the correct class is often ranked among the leading candidates even when it is not ranked first.

\begin{table}[h]
\caption{Zero-shot Top-$k$ classification accuracy (\%) on AID for the three decision functions, over all 10,000 images and 30 classes. An image is correct at Top-$k$ when its ground-truth class is among the $k$ highest-scoring concepts, and the Top-1 column equals the accuracies in \reftable{tab:aid_cross_model_comparison}. The best value in each column is in bold.}
\label{tab:aid_topk}
\centering
\begin{tabular}{|l|c|c|c|c|}
\hline
\textbf{Decision Function} & \textbf{Top-1} & \textbf{Top-3} & \textbf{Top-5} & \textbf{Top-10} \\
\hline
Presence Score \eqref{eq:decision_pres} & 37.26 & 57.05 & 66.44 & 79.79 \\
Conditional Localisation Score \eqref{eq:decision_logit} & 28.25 & 45.92 & 55.09 & 70.95 \\
Confidence Score \eqref{eq:confidence_score} & \textbf{38.70} & \textbf{57.57} & \textbf{66.44} & \textbf{79.82} \\
\hline
\end{tabular}
\end{table}

\subsubsection{Granular Per-Class Divergence: Vocabulary vs. Geometry}

\begin{figure*}[!t]
    \centering
    \includegraphics[width=\textwidth]{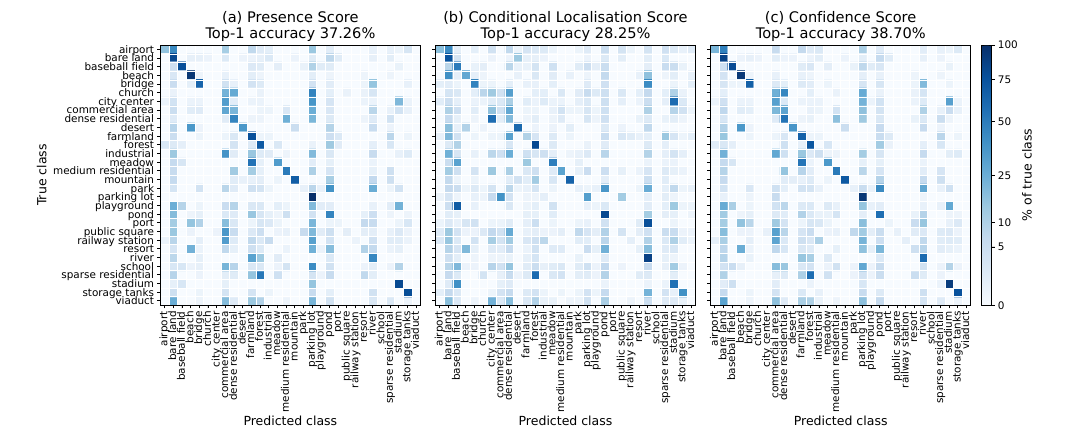}
    \caption{Top-1 confusion matrices on the AID dataset (10,000 images, 30 classes) for the three decision functions: (a) Presence Score, (b) Conditional Localisation Score, and (c) Confidence Score. Rows are the true class and columns the predicted class. Each cell shows the percentage of images of the true class that are assigned to the predicted class, so every row sums to 100\% and the diagonal gives the per-class Top-1 accuracy listed in \reftable{tab:appendix_aid_exhaustive}. The colour scale is square-root so that small off-diagonal rates remain visible.}
    \label{fig:aid_confusion}
\end{figure*}

To understand the mechanics of SAM~3's zero-shot classification, we analyse the quantitative class-wise performance detailed in \reftable{tab:appendix_aid_exhaustive} alongside the qualitative visual evidence presented in \reffig{fig:aid_qualitative} and the Top-1 confusion matrices in \reffig{fig:aid_confusion}. If we conceptually map this data, the target classes naturally partition into distinct operational clusters. These groupings illustrate how the inherent physical traits of a class govern which decision function ultimately succeeds or fails:

\begin{itemize}
    \item \textbf{Global Scene Dominance (Presence Wins):} When an entire scene inherently embodies a specific concept without relying on a single, isolated geometric boundary, the global presence token dominates the classification. For example, \textit{farmland} yields a strong Presence Score (77.03\%) but a poor Conditional Localisation Score (10.00\%). The \textit{farmland} row in \reffig{fig:aid_qualitative} demonstrates this perfectly: the Presence Score accurately captures the global scene context, whereas the Conditional Localisation Score gets distracted by localized sub-features, mistakenly highlighting patches of bare land or ponds in the examples shown. Across the whole class, the Localisation Score most often assigns \textit{farmland} images to dense residential, bare land, and sparse residential, as \reffig{fig:aid_confusion} shows.
    
    \item \textbf{Distinct Feature Geometry (Localisation Wins):} Conversely, when a target class is defined by large, clear, and easily isolatable physical features, localized geometric confidence drives the prediction. Natural amorphous features like \textit{rivers} yield a high Conditional Localisation Score (87.07\%) but a much lower Presence Score (44.15\%). As seen in the \textit{river} row of \reffig{fig:aid_qualitative}, the Conditional Localisation Score accurately traces the land-water boundaries to classify the image, while the global Presence Score misclassifies the broader surrounding scene as farmland. 
    
    \item \textbf{Synergistic Success (Both Win):} When a concept possesses both a massive, unambiguous geometric footprint and overwhelmingly dictates the global scene context, both heads succeed. The model performs exceptionally well on structures like \textit{parking lots} (100\% Presence, 93.33\% Confidence) and \textit{stadiums} (80.34\% Presence, 90.69\% Confidence). This synergy is visually confirmed in the \textit{stadium} examples in \reffig{fig:aid_qualitative}, where all three decision functions unanimously and correctly identify the target. At the class level, the Localisation Score alone is weaker on these classes (32.05\% for \textit{parking lots} and 55.52\% for \textit{stadiums}), so the examples shown are the cases where the two signals agree.
    
    \item \textbf{Cross-Modal Domain Failure (Both Fail):} Because SAM~3 is trained on a general-domain distribution, it severely struggles with concepts where it can neither reconcile the global scene nor isolate the defining local features from an overhead perspective. Accuracy falls to near zero for human-designated architectural properties like \textit{schools}, \textit{churches}, and \textit{city centers}, and Top-5 accuracy barely changes this picture under the Presence and Confidence Scores (0.42\% for \textit{church}, 0\% for \textit{school}, and 5.38\% and 6.54\% for \textit{city center}). The \textit{church} row in \reffig{fig:aid_qualitative} illustrates this complete semantic collapse, with all three decision functions failing entirely and hallucinating unrelated classes. Across the whole class, \reffig{fig:aid_confusion} shows that the predictions for church images concentrate on built-up classes such as dense residential, commercial area, and parking lot. A plausible explanation is a gap between the textual concept and its aerial appearance. A church appears as basic geometric rectangles from a nadir view, while the concept was likely learned mostly from ground-level views showing steeples and facades, so the top-down image may not match what the text prompt describes. We did not analyze the model's internal features, so this remains a hypothesis.
\end{itemize}

Top-5 accuracy also separates these classes from others that look similarly weak at Top-1. Under the Confidence Score, \textit{airport}, \textit{industrial}, and \textit{sparse residential} reach only 21.39\%, 4.36\%, and 0.33\% at Top-1 but 80.56\%, 74.36\%, and 63.67\% at Top-5, so the model ranks them among its leading candidates and another concept is simply scored higher on the same image. The confusion matrices in \reffig{fig:aid_confusion} show where such errors go. Under the Presence and Confidence Scores, bare land and parking lot receive a large share of the misclassified images, whereas under the Localisation Score the errors spread toward classes such as river, dense residential, and baseball field.

\begin{table}
\caption{Per-Class Scoring and Sample Breakdown on the AID Benchmark Dataset (10,000 images, 30 classes). The class concept in the first column is the textual prompt used, and Samples is the number of AID images of that class. Pres., Loc., and Conf. denote the zero-shot classification accuracy (\%) of the Presence Score, Conditional Localisation Score, and Confidence Score decision functions, each reported at Top-1 and Top-5. An image counts as correct at Top-$k$ when its ground-truth class is among the $k$ highest-scoring concepts.}
\label{tab:appendix_aid_exhaustive}
\centering
\resizebox{\columnwidth}{!}{
\begin{tabular}{|l|c|c|c|c|c|c|c|}
\hline
\multirow{2}{*}{\textbf{Class Concept}} & \multicolumn{2}{c|}{\textbf{Pres. Acc. (\%)}} & \multicolumn{2}{c|}{\textbf{Loc. Acc. (\%)}} & \multicolumn{2}{c|}{\textbf{Conf. Acc. (\%)}} & \multirow{2}{*}{\textbf{Samples}} \\
\cline{2-7}
 & Top-1 & Top-5 & Top-1 & Top-5 & Top-1 & Top-5 & \\
\hline
church              & 0 & 0.42 & 7.92 & 25.83 & 0 & 0.42 & 240 \\
stadium             & 80.34 & 99.66 & 55.52 & 90.34 & 90.69 & 99.31 & 290 \\
railway station     & 0.38 & 18.08 & 0.77 & 13.85 & 0.77 & 16.15 & 260 \\
farmland            & 77.03 & 88.11 & 10.00 & 47.30 & 65.95 & 89.19 & 370 \\
baseball field      & 77.73 & 98.64 & 50.45 & 81.82 & 79.55 & 98.64 & 220 \\
storage tanks       & 78.61 & 95.28 & 39.17 & 87.22 & 76.39 & 95.28 & 360 \\
beach               & 94.00 & 100 & 28.00 & 85.25 & 93.50 & 100 & 400 \\
parking lot         & 100 & 100 & 32.05 & 70.00 & 93.33 & 100 & 390 \\
airport             & 16.39 & 82.50 & 17.78 & 56.39 & 21.39 & 80.56 & 360 \\
forest              & 70.40 & 90.80 & 80.40 & 98.80 & 72.40 & 92.40 & 250 \\
park                & 0.57 & 10.00 & 0.57 & 10.86 & 0.57 & 9.43 & 350 \\
river               & 44.15 & 96.34 & 87.07 & 97.07 & 59.51 & 97.07 & 410 \\
school              & 0 & 0 & 0 & 0.67 & 0 & 0 & 300 \\
sparse residential  & 0 & 65.67 & 0 & 14.67 & 0.33 & 63.67 & 300 \\
pond                & 44.76 & 90.48 & 81.19 & 99.05 & 57.14 & 93.10 & 420 \\
bare land           & 81.61 & 99.03 & 74.84 & 94.84 & 86.13 & 99.35 & 310 \\
dense residential   & 45.12 & 99.02 & 19.76 & 74.15 & 54.39 & 99.02 & 410 \\
playground          & 1.08 & 8.65 & 0.27 & 31.08 & 0.54 & 9.46 & 370 \\
medium residential  & 50.00 & 98.28 & 30.00 & 77.24 & 50.69 & 98.97 & 290 \\
desert              & 35.67 & 89.33 & 60.67 & 98.67 & 37.00 & 92.00 & 300 \\
city center         & 0 & 5.38 & 1.15 & 19.23 & 0 & 6.54 & 260 \\
resort              & 9.66 & 67.24 & 0.69 & 10.00 & 4.14 & 61.38 & 290 \\
industrial          & 1.28 & 76.41 & 4.36 & 30.51 & 4.36 & 74.36 & 390 \\
public square       & 0.91 & 10.30 & 3.03 & 23.94 & 0.61 & 13.33 & 330 \\
viaduct             & 0 & 0 & 0 & 5.00 & 0 & 0 & 420 \\
mountain            & 67.35 & 92.35 & 65.29 & 90.00 & 70.00 & 93.24 & 340 \\
bridge              & 63.61 & 89.44 & 46.11 & 88.89 & 66.11 & 91.11 & 360 \\
port                & 0.26 & 5.53 & 0 & 2.11 & 0.26 & 3.42 & 380 \\
meadow              & 33.21 & 95.00 & 48.57 & 96.43 & 35.00 & 95.36 & 280 \\
commercial area     & 28.29 & 96.57 & 3.71 & 23.71 & 20.57 & 95.14 & 350 \\
\hline
\end{tabular}
}
\end{table}

\subsubsection{Cross-Model Comparison and Fine-Tuning Potential}
Comparing these results with other vision-language models in \reftable{tab:aid_cross_model_comparison} provides context regarding SAM~3's current limitations and downstream potential. The baseline accuracy for the OpenAI CLIP L/14 model is reported as benchmarked by Wang et al. \cite{wang2024skyscript}. The general-domain OpenAI CLIP \cite{radford2021clip} outpaces SAM~3 with a zero-shot accuracy of 69.25\%, as CLIP was engineered for global image-text contrastive learning to map holistic scene features directly to a text lexicon. Furthermore, when CLIP is fine-tuned on earth observation datasets, such as in GeoRSCLIP~\cite{georsclip2024} and RSDiX-CLIP~\cite{rsdixclip2025}, its scene classification accuracy reaches 73.72\% and 95.10\%, respectively.

This comparison illustrates both the current limitations and the potential of SAM~3. Currently, SAM~3 is constrained by its zero-shot vocabulary mapping, which aligns with physical ground-level object shapes rather than global aerial zoning. However, the class-level performance variations highlighted in the per-class analysis indicate that the model's visual segmentation infrastructure can successfully isolate aerial masks without prior domain training. Because the underlying spatial backbone is highly effective, a lightweight parameter-efficient fine-tuning stage or a domain-specific text projection layer could address these semantic misalignments, thereby enhancing performance in remote sensing applications.





\begin{figure*}[!t]
\centering
\subfloat[Isolated Airfield]{\includegraphics[width=0.48\textwidth]{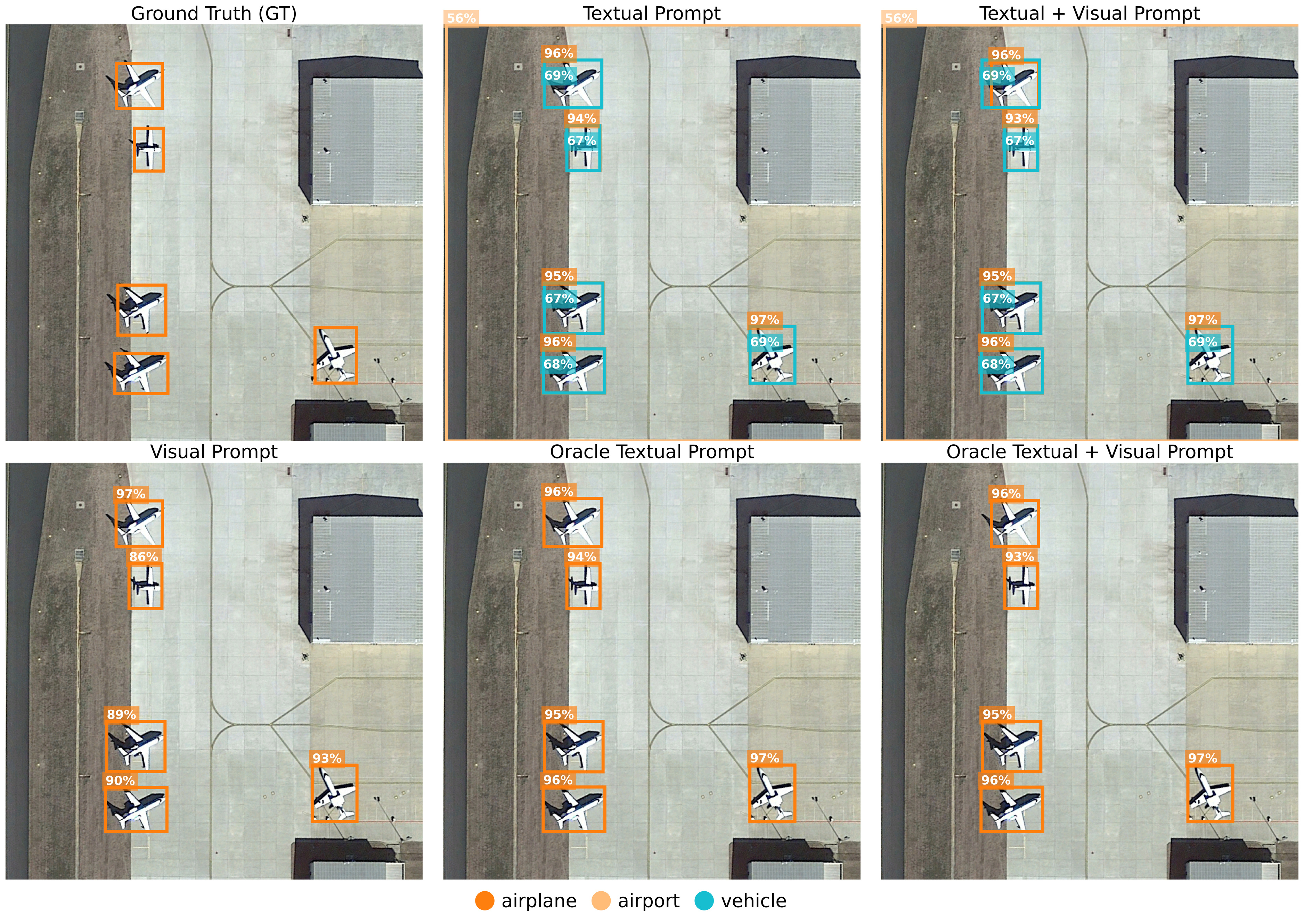}\label{fig:dior_plane1}}\hfil
\subfloat[Dense Complex Airfield]{\includegraphics[width=0.48\textwidth]{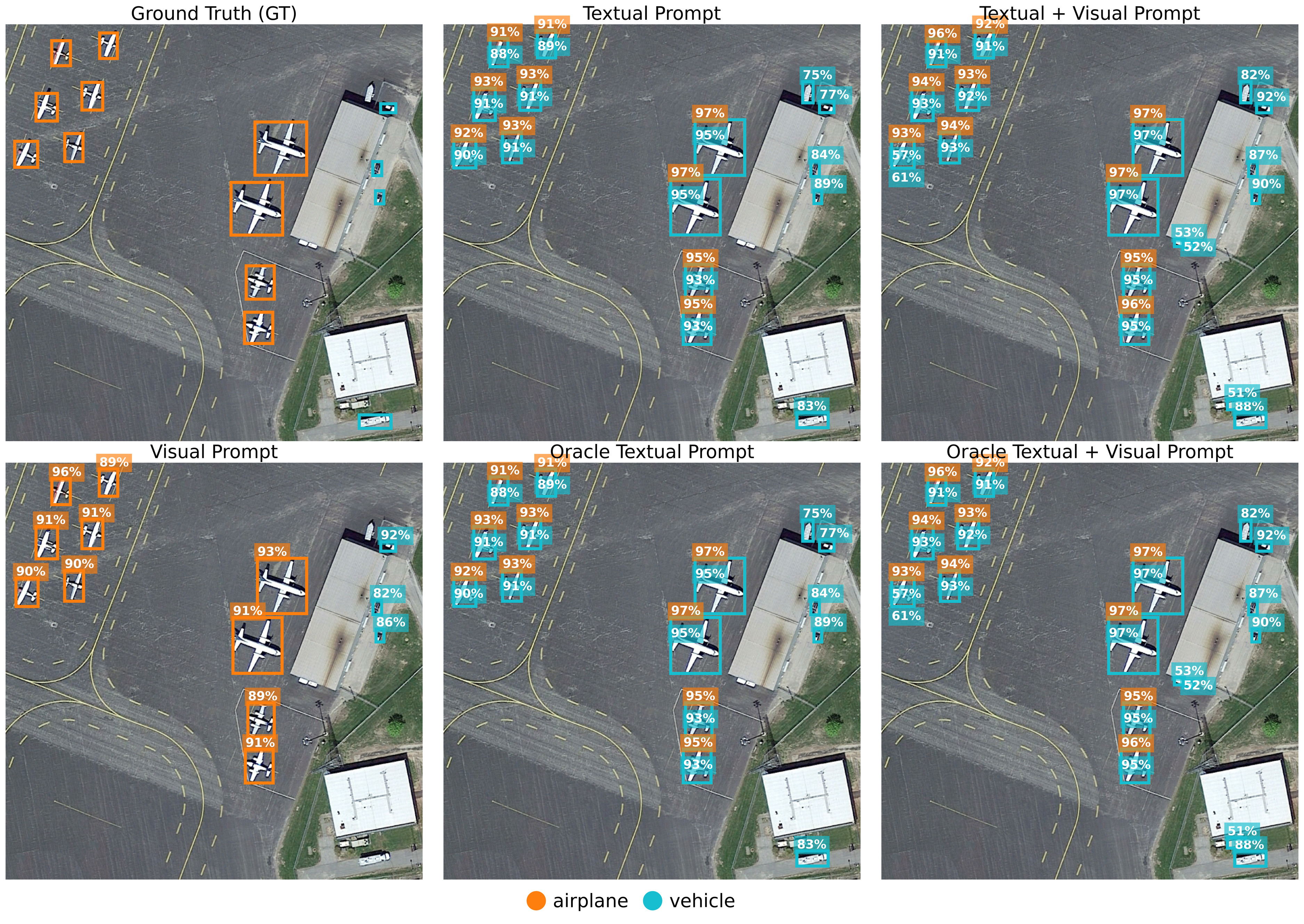}\label{fig:dior_plane2}}\\[0.4em]
\subfloat[Highway Overpass Intersection]{\includegraphics[width=0.48\textwidth]{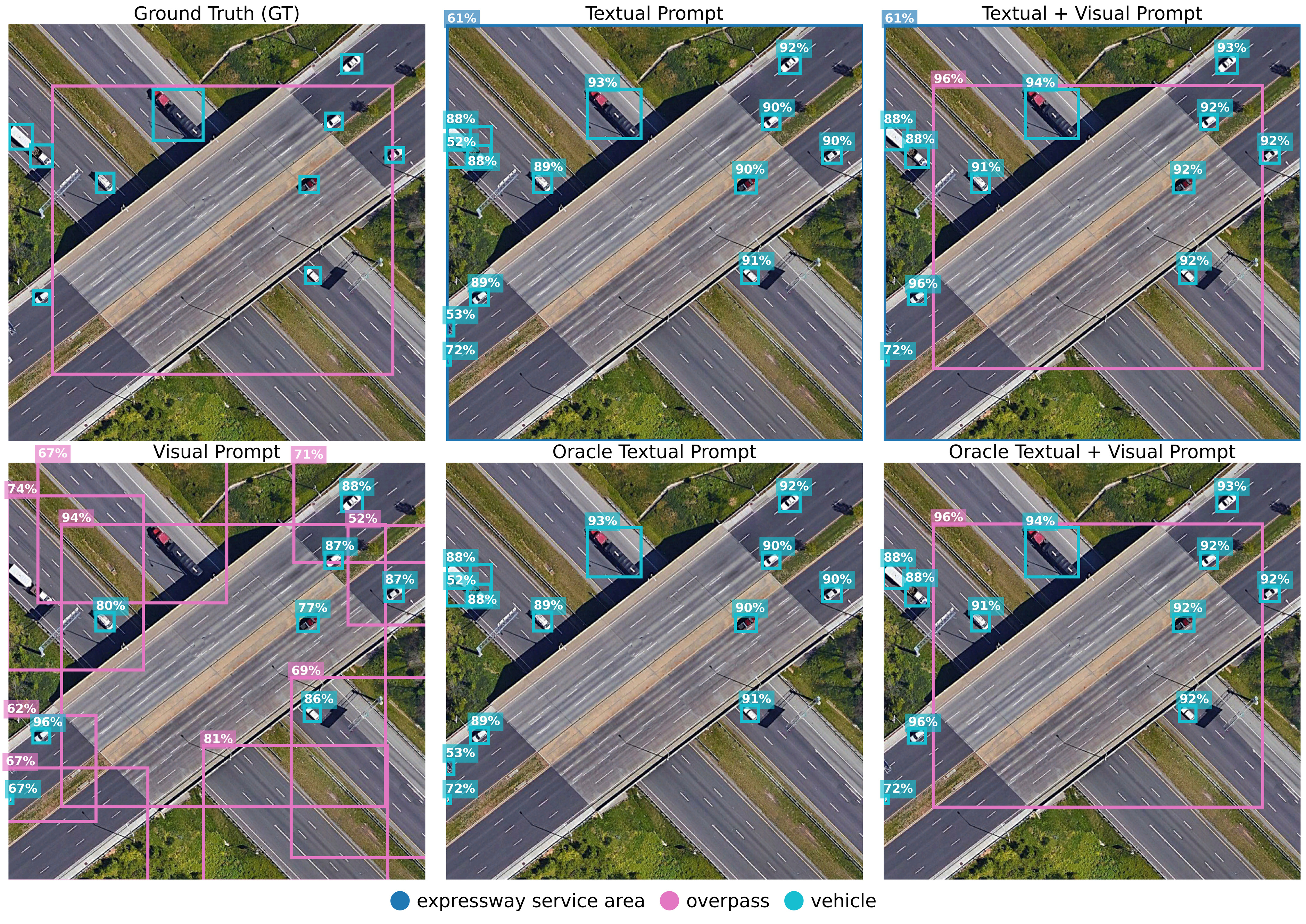}\label{fig:dior_overpass}}\hfil
\subfloat[Windmill Power Generation Field]{\includegraphics[width=0.48\textwidth]{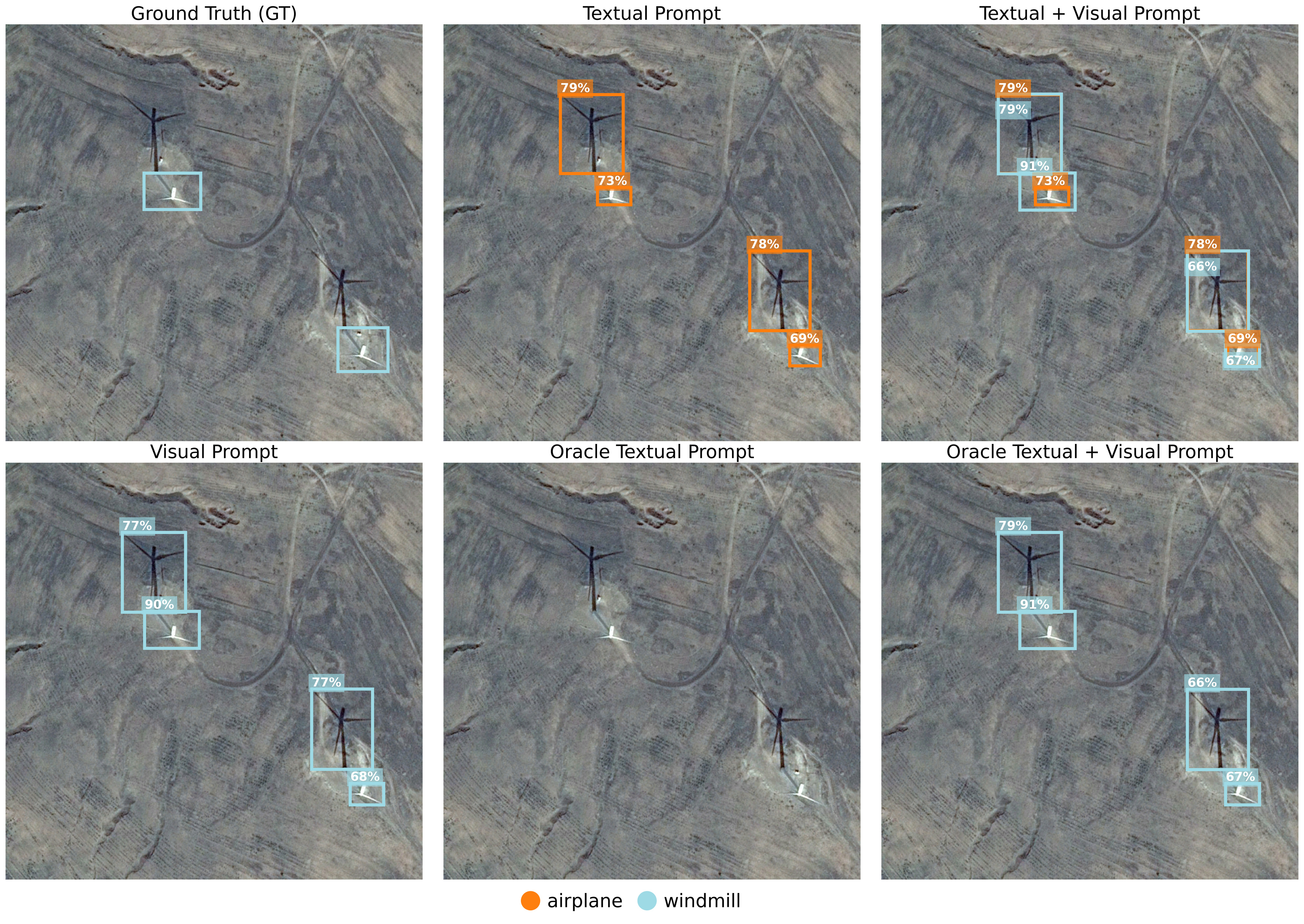}\label{fig:dior_windmill}}
\caption{Qualitative object detection results on the DIOR dataset across the five prompt configurations (\refConfs{config:text}{config:textbox_filt}). Each panel shows one DIOR image with the predictions of each configuration, labeled on its own sub-image, covering an isolated airfield, a dense complex airfield, a highway overpass intersection, and a windmill power generation field. Colors distinguish the different classes found in the image, and the ground truth is not shown.}
\label{fig:dior_qualitative}
\end{figure*}

\subsection{Open-Vocabulary Object Detection (DIOR)}
We evaluate SAM~3 within an open-vocabulary detection framework using the 20-class DIOR dataset. The performance metrics across our five core prompting configurations are summarized in \reftable{tab:dior_macro_ablation}. To observe the localized feature interactions underlying these scores, a qualitative prompt ablation across four distinct environments is presented in \reffig{fig:dior_qualitative}.

\begin{table*}[htbp]
\caption{Exhaustive Class-Wise Object Detection Evaluation on the 20 Classes of the DIOR Dataset. Each value is the Average Precision (\%) of a single class, reported as AP$_{50:95}$ (averaged over IoU thresholds 0.50 to 0.95) and AP$_{50}$ (at IoU 0.50). Modality Legend: T = Text-Only (\refConf{config:text}); TB = Text + Box (\refConf{config:textbox}); B = Box-Only (\refConf{config:box}); TF = Text + Oracle (\refConf{config:text_filt}); TBF = Text + Box + Oracle (\refConf{config:textbox_filt}). TB, B, and TBF columns report mean $\pm$ sample standard deviation over seven independently seeded random exemplar draws, while T and TF use no visual prompt and are deterministic, so they are reported from a single run.}
\label{tab:appendix_dior_merged}
\centering
\resizebox{\textwidth}{!}{
\begin{tabular}{|l|c|c|c|c|c|c|c|c|c|c|}
\hline
\multirow{2}{*}{\textbf{Class Concept}}& \multicolumn{5}{c|}{\textbf{AP$_{50:95}$}} & \multicolumn{5}{c|}{\textbf{AP$_{50}$}} \\
\cline{2-11}
 & \textbf{T} & \textbf{TB} & \textbf{B} & \textbf{TBF} & \textbf{TF} & \textbf{T} & \textbf{TB} & \textbf{B} & \textbf{TBF} & \textbf{TF} \\
\hline
Airplane & 60.22 & 64.05 $\pm$ 0.11 & 66.49 $\pm$ 0.21 & 65.99 $\pm$ 0.14 & 62.39 & 91.43 & 92.55 $\pm$ 0.36 & 95.64 $\pm$ 0.01 & 96.24 $\pm$ 0.50 & 95.61 \\
Airport & 0 & 92.34 $\pm$ 0.11 & 92.43 $\pm$ 0.05 & 92.44 $\pm$ 0.10 & 0 & 0 & 98.82 $\pm$ 0.14 & 98.91 $\pm$ 0.05 & 99.00 $\pm$ 0.01 & 0 \\
Baseball Field & 0.01 & 62.06 $\pm$ 0.39 & 78.32 $\pm$ 0.14 & 63.12 $\pm$ 0.40 & 0.01 & 0.01 & 69.41 $\pm$ 0.61 & 90.89 $\pm$ 0.40 & 70.78 $\pm$ 0.63 & 0.01 \\
Basketball Court & 3.05 & 70.96 $\pm$ 0.40 & 71.15 $\pm$ 0.20 & 71.78 $\pm$ 0.41 & 4.75 & 4.48 & 77.18 $\pm$ 0.46 & 76.62 $\pm$ 0.33 & 78.29 $\pm$ 0.47 & 6.92 \\
Bridge & 1.32 & 39.36 $\pm$ 0.19 & 39.03 $\pm$ 0.15 & 39.79 $\pm$ 0.20 & 1.58 & 3.14 & 51.51 $\pm$ 0.35 & 50.69 $\pm$ 0.22 & 52.31 $\pm$ 0.40 & 3.81 \\
Chimney & 0.80 & 73.46 $\pm$ 0.34 & 74.34 $\pm$ 0.32 & 73.51 $\pm$ 0.34 & 0.82 & 0.99 & 87.13 $\pm$ 0.72 & 87.55 $\pm$ 0.29 & 87.22 $\pm$ 0.73 & 0.99 \\
Dam & 2.79 & 82.33 $\pm$ 0.10 & 82.86 $\pm$ 0.11 & 82.34 $\pm$ 0.10 & 3.64 & 4.56 & 93.34 $\pm$ 0.46 & 93.84 $\pm$ 0.03 & 93.35 $\pm$ 0.48 & 6.08 \\
Expressway Service Area & 0.91 & 67.94 $\pm$ 0.54 & 71.38 $\pm$ 0.35 & 68.73 $\pm$ 0.54 & 2.20 & 1.03 & 82.75 $\pm$ 0.70 & 88.63 $\pm$ 0.54 & 84.01 $\pm$ 0.76 & 2.58 \\
Expressway Toll Station & 2.92 & 82.46 $\pm$ 0.09 & 84.05 $\pm$ 0.20 & 83.36 $\pm$ 0.13 & 8.37 & 4.92 & 91.93 $\pm$ 0.25 & 92.38 $\pm$ 0.22 & 93.37 $\pm$ 0.45 & 13.53 \\
Golf Field & 0 & 84.76 $\pm$ 0.32 & 87.50 $\pm$ 0.23 & 86.71 $\pm$ 0.33 & 0.35 & 0 & 90.45 $\pm$ 0.41 & 94.39 $\pm$ 0.32 & 93.12 $\pm$ 0.47 & 0.50 \\
Ground Track Field & 0.63 & 74.92 $\pm$ 0.23 & 79.60 $\pm$ 0.31 & 76.08 $\pm$ 0.26 & 1.78 & 1.20 & 79.49 $\pm$ 0.36 & 85.42 $\pm$ 0.45 & 81.08 $\pm$ 0.43 & 3.45 \\
Harbor & 18.82 & 54.52 $\pm$ 0.47 & 49.41 $\pm$ 1.10 & 54.99 $\pm$ 0.48 & 21.24 & 27.56 & 67.82 $\pm$ 0.54 & 62.57 $\pm$ 1.62 & 68.60 $\pm$ 0.55 & 31.27 \\
Overpass & 0.52 & 48.87 $\pm$ 0.12 & 50.83 $\pm$ 0.27 & 50.99 $\pm$ 0.13 & 2.02 & 1.21 & 62.14 $\pm$ 0.08 & 64.84 $\pm$ 0.44 & 65.73 $\pm$ 0.10 & 4.64 \\
Ship & 14.45 & 45.42 $\pm$ 0.49 & 50.33 $\pm$ 0.31 & 45.45 $\pm$ 0.49 & 14.49 & 21.89 & 69.14 $\pm$ 0.70 & 76.66 $\pm$ 0.52 & 69.18 $\pm$ 0.70 & 21.96 \\
Stadium & 44.69 & 91.96 $\pm$ 0.40 & 92.10 $\pm$ 0.38 & 92.20 $\pm$ 0.42 & 49.82 & 55.70 & 95.52 $\pm$ 0.47 & 95.62 $\pm$ 0.56 & 95.82 $\pm$ 0.52 & 62.42 \\
Storage Tank & 25.10 & 49.51 $\pm$ 0.16 & 49.19 $\pm$ 0.33 & 49.75 $\pm$ 0.16 & 25.88 & 38.20 & 74.57 $\pm$ 0.03 & 73.48 $\pm$ 0.52 & 75.02 $\pm$ 0.02 & 39.46 \\
Tennis Court & 38.33 & 77.94 $\pm$ 0.16 & 78.85 $\pm$ 0.13 & 78.20 $\pm$ 0.17 & 40.10 & 45.47 & 89.61 $\pm$ 0.40 & 89.87 $\pm$ 0.31 & 90.02 $\pm$ 0.41 & 47.61 \\
Train Station & 0.63 & 85.44 $\pm$ 0.09 & 84.28 $\pm$ 0.13 & 85.44 $\pm$ 0.09 & 0.63 & 0.99 & 97.92 $\pm$ 0.00 & 97.36 $\pm$ 0.20 & 97.92 $\pm$ 0.00 & 0.99 \\
Vehicle & 17.55 & 24.71 $\pm$ 0.05 & 33.91 $\pm$ 0.33 & 28.46 $\pm$ 0.06 & 21.47 & 32.00 & 39.31 $\pm$ 0.15 & 57.78 $\pm$ 0.84 & 48.19 $\pm$ 0.20 & 41.41 \\
Windmill & 0.74 & 28.61 $\pm$ 0.13 & 30.86 $\pm$ 0.16 & 28.62 $\pm$ 0.13 & 0.74 & 0.99 & 39.62 $\pm$ 0.22 & 45.26 $\pm$ 0.30 & 39.64 $\pm$ 0.22 & 0.99 \\
\hline
\end{tabular}
}
\end{table*}

\begin{table}
\caption{Object detection performance on DIOR, reported as mAP$_{50:95}$ and mAP$_{50}$ (Average Precision averaged across all classes, at IoU 0.50 to 0.95 and at IoU 0.50 respectively), across the five prompt configurations defined in \sectionref{subsec:prompt_modalities}: Text-Only, Box-Only, Text + Box, and their ground-truth oracle-filtered variants. Box-Only, Text + Box, and Text + Box + Oracle are reported as mean $\pm$ sample standard deviation over seven independently seeded random exemplar draws, while Text-Only and Text + Oracle use no visual prompt and are deterministic, so they are reported from a single run. The best value in each column is in bold.}
\label{tab:dior_macro_ablation}
\centering
\begin{tabular}{|l|c|c|}
\hline
\textbf{Setup} & \textbf{mAP$_{50:95}$} & \textbf{mAP$_{50}$} \\
\hline
Text-Only (\refConf{config:text})    & 11.67 & 16.79 \\
Box-Only (\refConf{config:box})     & \textbf{67.35 $\pm$ 0.09} & \textbf{80.92 $\pm$ 0.12} \\
Text + Box (\refConf{config:textbox})  & 65.08 $\pm$ 0.04 & 77.51 $\pm$ 0.10 \\
Text + Oracle (\refConf{config:text_filt})  & 13.11 & 19.21 \\
Text + Box + Oracle (\refConf{config:textbox_filt}) & 65.90 $\pm$ 0.05 & 78.94 $\pm$ 0.12 \\
\hline
\end{tabular}
\end{table}

\subsubsection{Cross-Modal Alignment Gap and Linguistic Interference}
Analyzing the experimental configuration reveals a severe polarization between SAM~3's textual alignment and its pure visual parsing abilities. The zero-shot baseline using text prompts (\refConf{config:text}) yields an 11.67 mAP$_{50:95}$. Conversely, introducing visual guidance through a single bounding box exemplar (\refConf{config:box}) forces an absolute localization surge to 67.35 $\pm$ 0.09 mAP$_{50:95}$ (80.92 $\pm$ 0.12 mAP$_{50}$). This indicates that SAM~3 handles dense remote sensing scenes well when matching a visual prompt to the image, although each score includes the prompted instance itself.

Comparing Box-Only against Text + Box mixes two different things, since Box-Only only ever answers for present classes while Text + Box also answers for absent ones with text alone. Staying inside the group that only scores present classes, adding text to a box lowers aggregate localization from 67.35 mAP$_{50:95}$ under Box-Only to 65.90 mAP$_{50:95}$ under Text + Box + Oracle, a drop of 1.45 points (\reftable{tab:cross_modal_effect}). The 95\% bootstrap interval of this change excludes zero, so the drop is not explained by which images happen to be in the test set. The class-wise changes (\reftable{tab:cross_modal_per_class}) show that it is also uneven. Only 8 of the 20 classes improve, and the drop is concentrated in a few classes, most of all baseball field ($-15.20$), followed by vehicle and ship, while harbor gains the most ($+5.58$). The plain Text + Box result, 65.08 mAP$_{50:95}$, drops further still, to 2.27 points below Box-Only, and that extra 0.82-point drop comes from false positives on classes that are not actually present. This remaining gap is illustrated by the overpass detection task in \reffig{fig:dior_overpass}. Observing the image, the pure visual prompt fails to isolate the overpass, instead scattering bounding boxes across the connected highway system, while the pure text prompt yields zero detections. However, when both prompts are mixed, the model accurately isolates the specific boundaries of the overpass.

A plausible explanation is that text can act as a helpful signal here precisely because the visual prompt alone is ambiguous. An overpass may look visually close to a standard road from directly overhead, which could explain why the visual-only mode over-segments the surrounding highway. The word ``overpass'' may carry a more specific, road-adjacent structure association that helps narrow the region down once combined with the visual prompt, correcting the final bounding boxes. We did not analyze the model's internal representations to confirm this.

\subsubsection{Presence Head Filtering Analysis}
Since the oracle configurations use ground-truth class presence, they show what a perfect presence predictor could add on top of the model's own detections, not a result the model could reach on its own. Moving from \refConf{config:text} (11.67 mAP$_{50:95}$) to \refConf{config:text_filt} (13.11 mAP$_{50:95}$) and from \refConf{config:textbox} (65.08 mAP$_{50:95}$) to \refConf{config:textbox_filt} (65.90 mAP$_{50:95}$) yields a consistent but modest increase. This change suggests that without active prompting suppression, the model experiences false-positive detections over structurally similar elements.

This behavior is illustrated in the windmill field example in \reffig{fig:dior_windmill}. In the unfiltered text and multimodal configurations, the model misclassifies the windmills, generating false-positive ``airplane'' bounding boxes around them. When the negative oracle filter is applied, these false-positive airplane boxes are suppressed. 

A plausible mechanism is that the top-down shape of a windmill's blades resembles the wings and fuselage of an aircraft closely enough that, given the text-vision alignment was likely learned mostly from ground-level imagery, the decoder's search shifts toward the \textit{airplane} representation instead of \textit{windmill}. The oracle filter counteracts this by removing the mismatched airplane queries from the downstream decoders. This suggests that while the decoupled presence head introduces background noise, it is not the primary performance bottleneck, and that the main operational constraint remains false negatives, consistent with the difficulty of mapping semantic text prompts to unfamiliar top-down aerial geometries. We did not analyze the model's internal features to confirm this interpretation.

\subsubsection{Granular Class Analysis: Persistent Vulnerabilities and Structural Drops}
\label{subsubsec:dior_granular}

A class-by-class analysis of the performance matrix in \reftable{tab:appendix_dior_merged} and the visualisation in \reffig{fig:dior_qualitative} reveals localized variations that persist across the five evaluation configurations:

\begin{itemize}

\item \textbf{Semantic Performance on Macro Zones:} Large-scale spatial categories completely collapse under the pure zero-shot text Configuration (\refConf{config:text}). \textit{Airport} drops to an absolute 0~AP$_{50:95}$ in both the Text-Only (\refConf{config:text}) and Text + Oracle (\refConf{config:text_filt}) configurations, but immediately shoots up to 92.43~AP$_{50:95}$ under the visual prompt configuration (\refConf{config:box}) and 92.44~AP$_{50:95}$ under the multimodal configuration with oracle filtering (\refConf{config:textbox_filt}). Similarly, \textit{Baseball Field} sits at a non-functional 0.01~AP$_{50:95}$ under the pure text modality, peaks at 78.32~AP$_{50:95}$ under the visual prompt configuration, and suffers a massive drop back down to 63.12~AP$_{50:95}$ under the multimodal configuration with oracle filtering. A plausible explanation is that these sprawling, heterogeneous regions lack a distinct, localized visual signature for a text prompt to anchor onto. When textual and visual prompts are fused (\refConf{config:textbox}), the text prompt may act as noise on top of the visual signal, pulling the prediction away from its target and contributing to the sharp drop observed in categories like the baseball field. We did not analyze the model's internal representations to confirm this.

\item \textbf{Text-Induced Clutter and Vehicle Interaction:} Unlike large zones, distinct geometric targets show strong initial text affinity, with \textit{Airplane} scoring a high 60.22~AP$_{50:95}$ under the pure zero-shot text Configuration (\refConf{config:text}). However, unguided text embeddings trigger heavy background clutter. Observing the airplane task in \reffig{fig:dior_plane1}, the text prompt successfully captures the airplanes but hallucinates false-positive boxes over background vehicles that are not labeled in the ground truth. Applying the oracle filtering (\refConf{config:text_filt}) suppresses these errors. Conversely, in \reffig{fig:dior_plane2}, where vehicles are actively present in the ground truth, the oracle filtering correctly leaves these vehicle boxes intact. This is consistent with \reftable{tab:appendix_dior_merged}, where \textit{Vehicle} accuracy rises from 17.55~AP$_{50:95}$ under the Text-Only configuration to 21.47~AP$_{50:95}$ under the Text + Oracle configuration, suggesting that the text prompt maps reasonably well to distinct shapes but benefits from visual prompting or oracle filtering to prevent background spillover.

\item \textbf{The Shadow Distortion Constraint:} A plausible physical bottleneck is the model's difficulty decoupling an object's structural boundary from its projected shadow, visible across both the windmill blades in \reffig{fig:dior_windmill} and the vehicles in \reffig{fig:dior_plane2}. This is consistent with \reftable{tab:appendix_dior_merged}, where \textit{Windmill} stalls at a low 30.86~AP$_{50:95}$ even under the visual prompt configuration (\refConf{config:box}), and \textit{Vehicle} drops from 33.91~AP$_{50:95}$ under the visual prompt configuration to 28.46~AP$_{50:95}$ under the multimodal configuration with oracle filtering (\refConf{config:textbox_filt}). One possible explanation is that general-domain ground imagery rarely contains prominent, top-down projected shadows, so this visual pattern may be unfamiliar to the model and could contribute to the coordinate regression jitter observed in these classes. We did not analyze the model's internal features to confirm this.

\end{itemize}

\begin{table}[h]
\caption{Size-stratified detection AP for selected DIOR classes, comparing the weakest configuration (Text-Only) against the strongest (Text + Box + Oracle). AP$_S$, AP$_M$, AP$_L$ denote AP for small, medium, and large objects by pixel area, following the COCO definition. The All classes row reports the mean across classes, mAP$_S$, mAP$_M$, and mAP$_L$. Text + Box + Oracle values are reported as mean $\pm$ sample standard deviation over seven independently seeded random exemplar draws.}
\label{tab:dior_size_stratified}
\centering
\resizebox{\columnwidth}{!}{
\begin{tabular}{|l|c|c|c|c|c|c|}
\hline
\multirow{2}{*}{\textbf{Class}} & \multicolumn{3}{c|}{\textbf{Text-Only}} & \multicolumn{3}{c|}{\textbf{Text + Box + Oracle}} \\
\cline{2-7}
 & AP$_S$ & AP$_M$ & AP$_L$ & AP$_S$ & AP$_M$ & AP$_L$ \\
\hline
All classes & 5.07 & 13.74 & 18.82 & 40.29 $\pm$ 0.69 & 64.44 $\pm$ 0.21 & 80.94 $\pm$ 0.28 \\
Airport & -- & 0.00 & 0.00 & -- & 82.37 $\pm$ 1.21 & 92.61 $\pm$ 0.04 \\
Baseball Field & 0.00 & 0.18 & 0.01 & 48.12 $\pm$ 0.97 & 57.59 $\pm$ 0.93 & 74.37 $\pm$ 0.36 \\
Airplane & 44.00 & 65.15 & 77.92 & 51.43 $\pm$ 0.22 & 70.03 $\pm$ 0.15 & 83.68 $\pm$ 0.24 \\
Vehicle & 12.08 & 51.19 & 22.42 & 24.73 $\pm$ 0.11 & 62.37 $\pm$ 0.15 & 49.63 $\pm$ 5.17 \\
Windmill & 0.00 & 0.74 & 0.00 & 14.72 $\pm$ 0.38 & 36.53 $\pm$ 0.33 & 74.76 $\pm$ 0.43 \\
\hline
\end{tabular}
}
\end{table}

Table~\ref{tab:dior_size_stratified} supports this pattern with a direct size stratification. Across all classes, mAP$_S$ trails mAP$_M$ and mAP$_L$ under every configuration, for example 40.29 versus 64.44 and 80.94 under Text + Box + Oracle. Windmill shows this clearly at the class level. Its AP$_L$ reaches 74.76, while its AP$_S$ and AP$_M$ remain far lower, consistent with the shadow and resolution related explanation discussed in \sectionref{subsubsec:dior_granular} for this class. We did not repeat inference at multiple input resolutions, so this remains supporting evidence rather than a direct test of the resolution hypothesis. Vehicle's AP$_L$ also carries a notably wider spread across exemplar draws than any other cell in this table (49.63 $\pm$ 5.17), which we attribute to Vehicle's large-instance ground truth being both sparse and visually heterogeneous, making the single randomly drawn exemplar more consequential for this class and size bucket than for the others.

\subsubsection{Comparison Against General-Domain Open-Vocabulary Detectors}
\reftable{tab:ovd_baseline_comparison} places SAM~3's zero-shot text performance alongside four general-domain, open-vocabulary detection models under the protocol described in Section~\ref{subsec:eval_protocol}.

\begin{table}[t]
\caption{Zero-shot detection (DIOR, 20 classes) and instance segmentation (iSAID, 15 classes) performance of SAM~3 against general-domain, open-vocabulary baselines, reported as mAP$_{50}$ and mAP$_{50:95}$ (\%). All models are prompted with the dataset class names only. SAM~3 is the Text-Only configuration (\refConf{config:text}) and SAM~3 (Oracle) is the Text + Oracle configuration (\refConf{config:text_filt}). gDINO-T and gDINO-B denote Grounding DINO Tiny and Base, S and L denote the small and large variants of YOLO-World and YOLOE, and * denotes the ensemble variant. \dag\ denotes segmentation obtained via SAM~1 box-to-mask conversion, and unmarked segmentation values are produced natively by the model itself. A dash marks a value that was not computed.}
\label{tab:ovd_baseline_comparison}
\centering
\renewcommand{\arraystretch}{1.15}
\begin{tabular}{|l|c|c|c|c|}
\hline
\multirow{2}{*}{\textbf{Model}} & \multicolumn{2}{c|}{\makecell{\textbf{Box}\\\textbf{Detection}}} & \multicolumn{2}{c|}{\makecell{\textbf{Instance}\\\textbf{Segmentation}}} \\
\cline{2-5}
 & \textbf{mAP$_{50}$} & \textbf{mAP$_{50:95}$} & \textbf{mAP$_{50}$} & \textbf{mAP$_{50:95}$} \\
\hline
OWLv2~\cite{minderer2023owlv2}\dag & 12.63 & 7.05 & 10.31 & 4.36 \\
OWLv2*~\cite{minderer2023owlv2}\dag & 15.92 & 9.40 & 16.14 & 7.20 \\
\hline
gDINO-T~\cite{liu2023grounding}\dag & 2.65 & 1.32 & 3.41 & 1.06 \\
gDINO-B~\cite{liu2023grounding}\dag & 4.25 & 2.31 & 4.93 & 1.61 \\
\hline
YOLO-World-S~\cite{yoloworld2024}\dag & 1.54 & 0.76 & 1.49 & 0.46 \\
YOLO-World-L~\cite{yoloworld2024}\dag & 2.21 & 1.18 & 2.23 & 0.67 \\
\hline
YOLOE-S~\cite{wang2025yoloe}\dag & 1.68 & 0.89 & 1.91 & 0.54 \\
YOLOE-S~\cite{wang2025yoloe} & -- & -- & 1.52 & 0.32 \\
YOLOE-L~\cite{wang2025yoloe}\dag & 2.43 & 1.33 & 3.03 & 0.93 \\
YOLOE-L~\cite{wang2025yoloe} & -- & -- & 2.84 & 0.68 \\
\hline
\textbf{SAM~3} & \textbf{16.79} & \textbf{11.67} & \textbf{22.73} & \textbf{11.85} \\
\textbf{SAM~3 (Oracle)} & 19.21 & 13.11 & 24.15 & 12.70 \\
\hline
\end{tabular}
\end{table}

SAM~3 leads every baseline on DIOR, beating the strongest one, OWLv2's ensemble variant, by 0.87 mAP$_{50}$. The gap is more telling under the stricter mAP$_{50:95}$ metric, where SAM~3 leads by 2.27 points, a relative advantage more than four times larger than under mAP$_{50}$. The two models are not simply finding a similar number of correct detections. SAM~3's boxes are more precisely localized once partial overlaps are penalized.

Within every model family the larger checkpoint outperforms the smaller one, OWLv2's ensemble variant over its base variant, Grounding DINO's base over tiny, and YOLO-World's large over small, a consistent pattern across independent architectures that supports the reliability of the comparison itself. Grounding DINO and YOLO-World trail OWLv2 by a wide margin, roughly an order of magnitude at mAP$_{50}$ rather than a few points. One plausible contributor is training data composition. YOLO-World is trained exclusively on ground-level photography (Objects365, GQA, Flickr30k, CC3M), with no aerial or drone imagery of any kind, while OWLv2's self-training recipe draws on a much larger, less curated pool of web images that may incidentally include some overhead perspectives. This would not make OWLv2 a remote sensing model in any meaningful sense, but it may explain why its transfer to DIOR is comparatively less catastrophic than the other three.

Applying ground-truth presence filtering to SAM~3's text-only predictions raises its mAP$_{50}$ from 16.79 to 19.21, a modest gain next to the far larger jump that visual exemplar guidance produced in \reftable{tab:dior_macro_ablation}. Read together, these results suggest that SAM~3's advantage over other zero-shot, text-only detectors is real but narrow, and that text-only cross-modal transfer remains the binding constraint for all five models tested here, SAM~3 included. The much stronger numbers achieved once a visual exemplar is introduced show that the underlying localization capability is not the bottleneck. The gap between what text alone can trigger and what the model is actually capable of finding is the more important finding of this comparison.





\begin{figure*}[!t]
\centering
\subfloat[Coastal Infrastructure]{\includegraphics[width=0.48\textwidth]{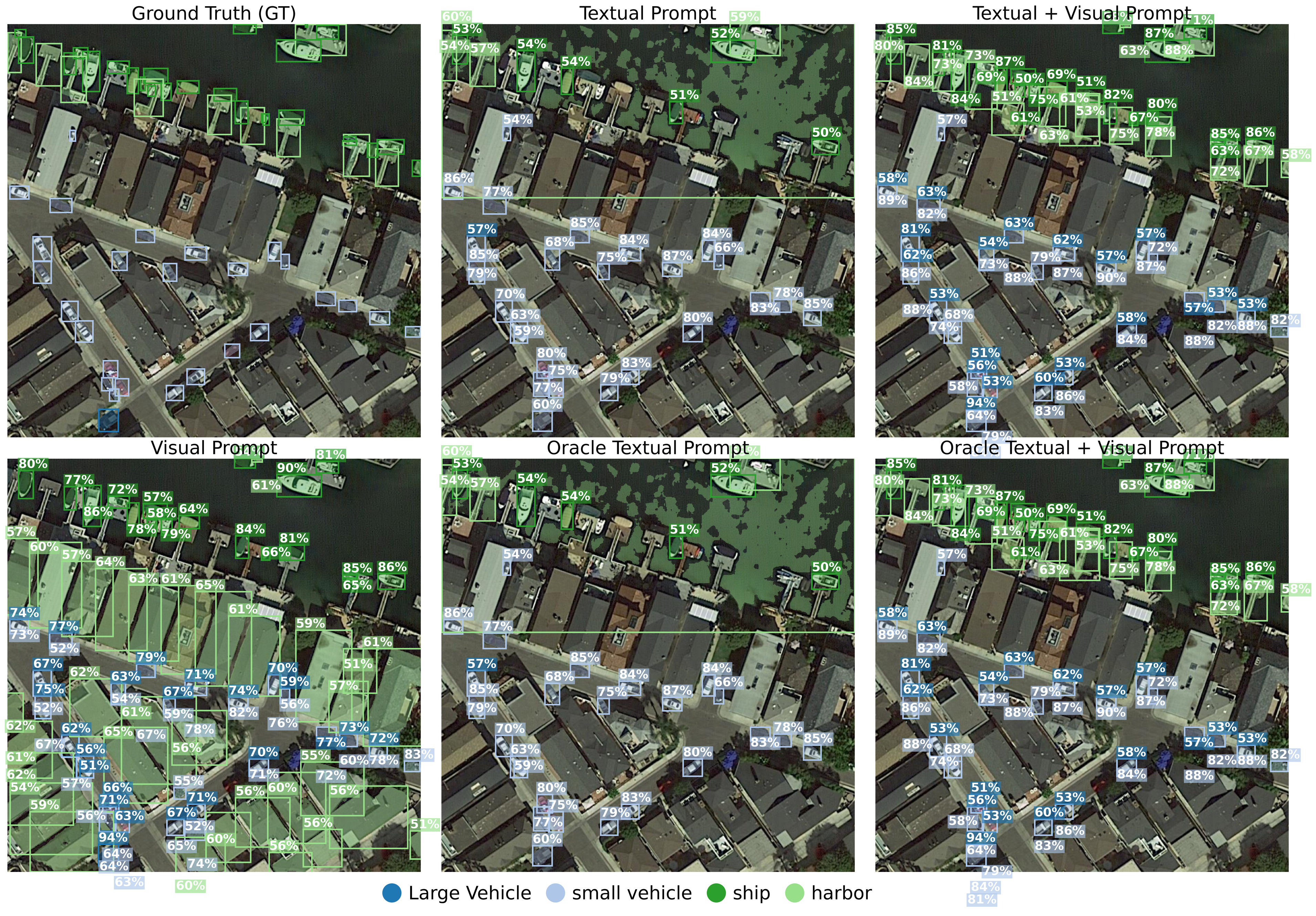}\label{fig:isaid_harbor}}\hfil
\subfloat[Sports Fields]{\includegraphics[width=0.48\textwidth]{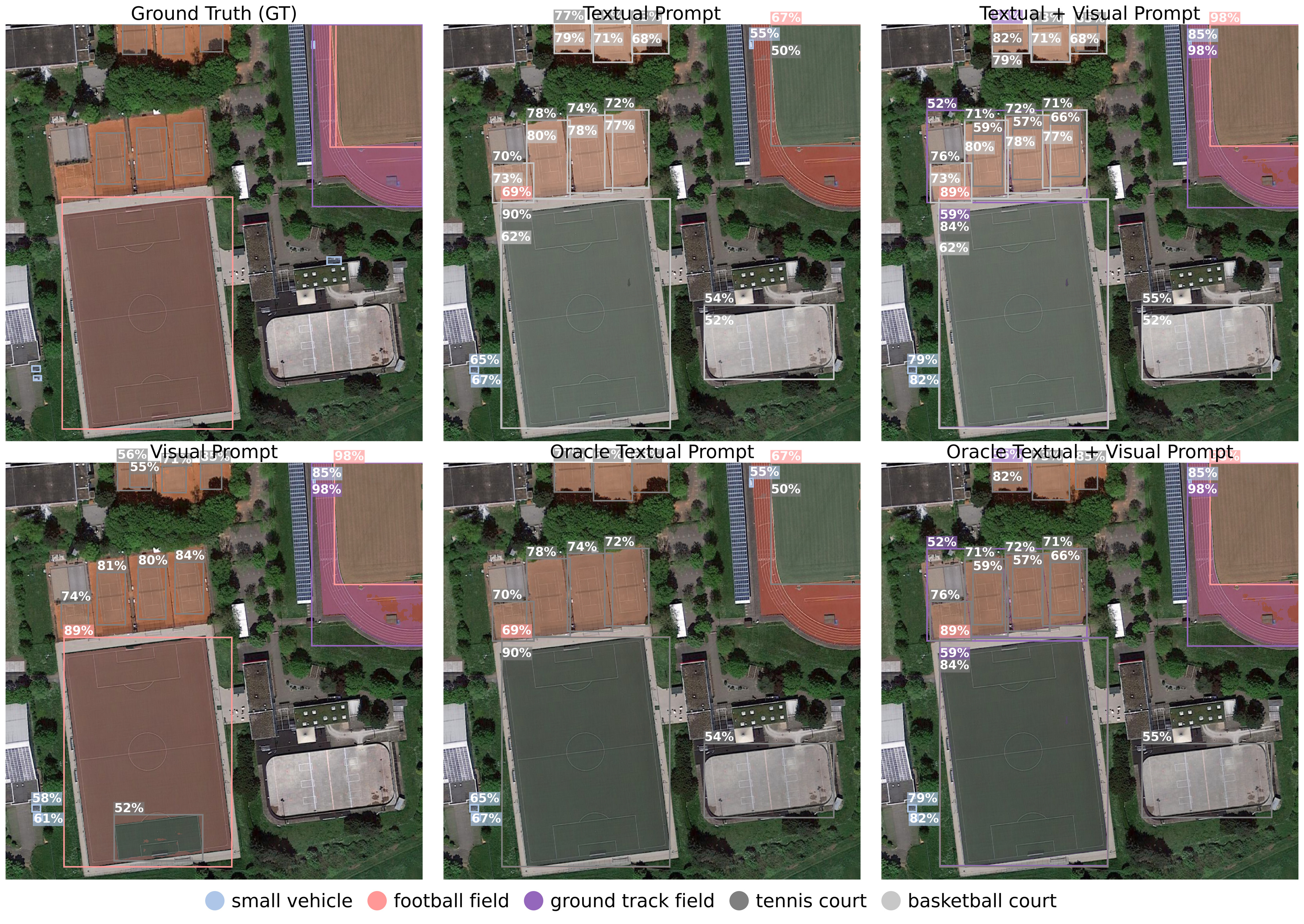}\label{fig:isaid_sports}}\\[0.4em]
\subfloat[Baseball Diamond]{\includegraphics[width=0.48\textwidth]{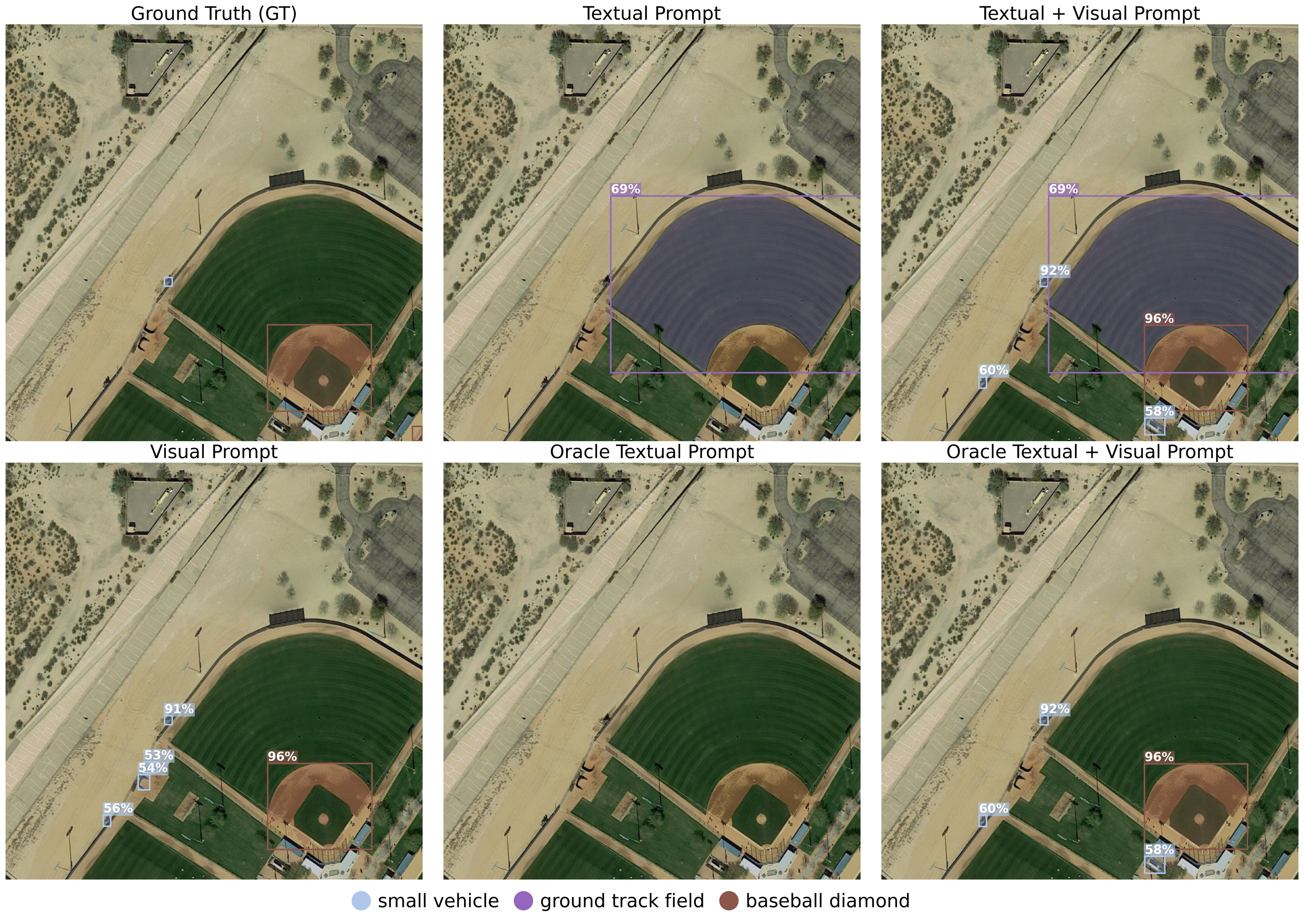}\label{fig:isaid_baseball}}\hfil
\subfloat[Roundabout]{\includegraphics[width=0.48\textwidth]{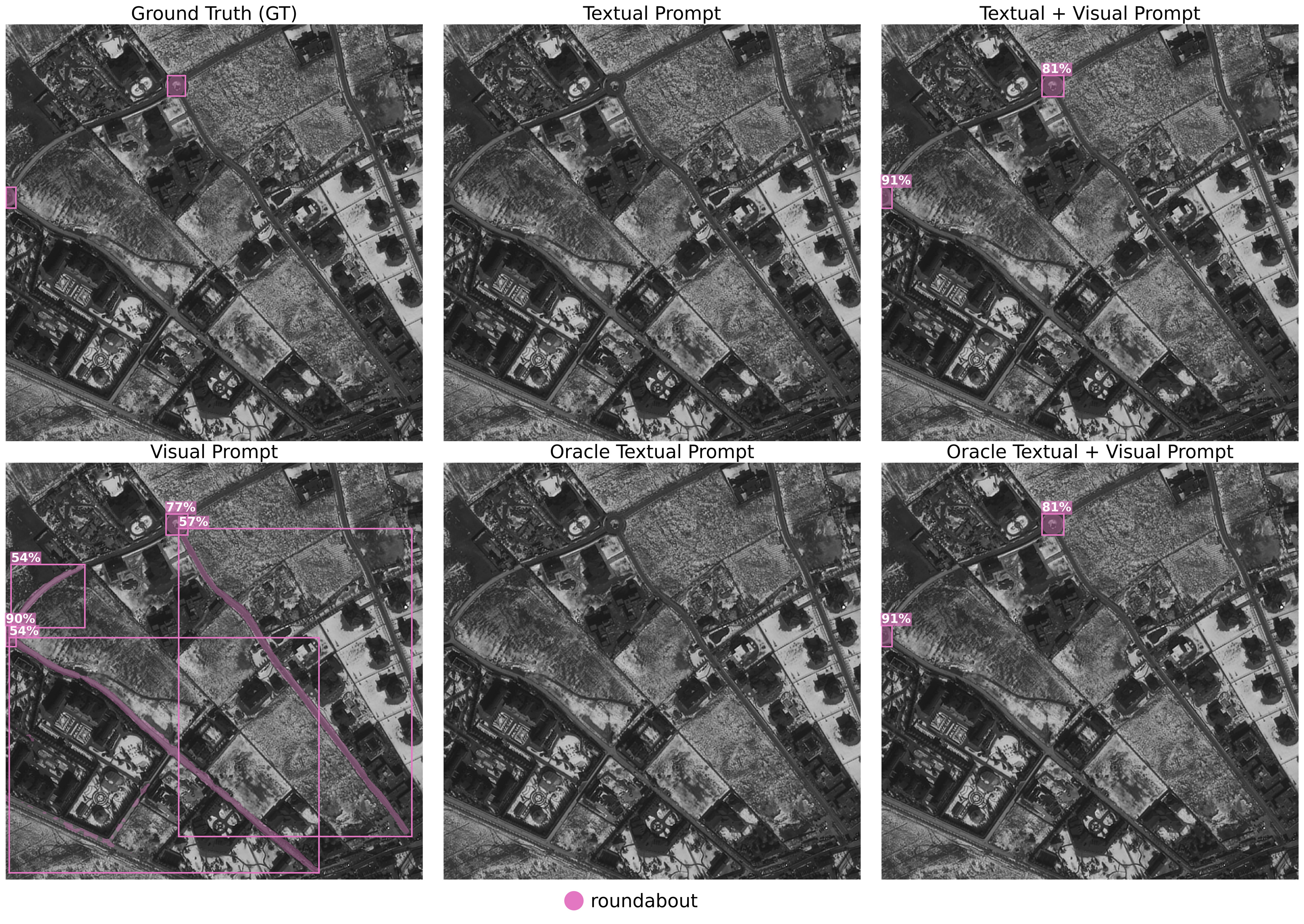}\label{fig:isaid_roundabout}}
\caption{Qualitative instance segmentation results on the iSAID dataset across the prompt configurations (\refConfs{config:text}{config:textbox_filt}). Each panel shows one iSAID image patch with the predictions of each configuration, labeled on its own sub-image, covering coastal infrastructure, sports fields, a baseball diamond, and a roundabout. Colors distinguish the different classes found in the image patch, and the ground truth is not shown. The panels illustrate how spatial scale, visual ambiguity, and textual perspective gaps can affect the success or failure of the multimodal decoder.}
\label{fig:isaid_qualitative}
\end{figure*}

\subsection{Open-Vocabulary Instance Segmentation (iSAID)}
Finally, we evaluate SAM~3's Promptable Concept Segmentation engine on the 15-class iSAID dataset. The macro performance profiles across the five functional modalities are summarized in \reftable{tab:isaid_macro_ablation}. To mechanistically contextualize these quantitative metrics, a qualitative prompt ablation across four distinct spatial environments is presented in \reffig{fig:isaid_qualitative}.

\begin{table}[h]
\caption{Instance segmentation performance on iSAID, reported as mAP$_{50:95}$ and mAP$_{50}$ (Average Precision averaged across all classes, at IoU 0.50 to 0.95 and at IoU 0.50 respectively), across the same five prompt configurations used for DIOR (\sectionref{subsec:prompt_modalities}). Box-Only, Text + Box, and Text + Box + Oracle are reported as mean $\pm$ sample standard deviation over seven independently seeded random exemplar draws, while Text-Only and Text + Oracle use no visual prompt and are deterministic, so they are reported from a single run. The best value in each column is in bold.}
\label{tab:isaid_macro_ablation}
\centering
\begin{tabular}{|l|c|c|}
\hline
\textbf{Setup} & \textbf{mAP$_{50:95}$} & \textbf{mAP$_{50}$} \\
\hline
Text-Only (\refConf{config:text})    & 11.85 & 22.73 \\
Box-Only (\refConf{config:box})     & 32.81 $\pm$ 0.16 & 57.18 $\pm$ 0.19 \\
Text + Box (\refConf{config:textbox})  & 35.35 $\pm$ 0.12 & 61.05 $\pm$ 0.22 \\
Text + Oracle (\refConf{config:text_filt})  & 12.70 & 24.15 \\
Text + Box + Oracle (\refConf{config:textbox_filt}) & \textbf{36.07 $\pm$ 0.13} & \textbf{62.43 $\pm$ 0.23} \\
\hline
\end{tabular}
\end{table}

\subsubsection{Text-Only Baseline Deficit vs. Visual-Prompt Calibration}

Ablating the prompting modalities on the iSAID dataset reveals a severe baseline performance deficit when relying strictly on textual guidance, with the Text-Only Configuration (\refConf{config:text}) bottlenecking at a low 11.85 mAP$_{50:95}$ and 22.73 mAP$_{50}$. Observing the visual outputs, the model completely fails to segment targets like roundabouts, as shown in \reffig{fig:isaid_roundabout}) and baseball diamonds, as shown in \reffig{fig:isaid_baseball}, when given only a text prompt. 

This total absence of detection is consistent with a large gap between the textual concept and its aerial appearance, although we did not analyze the model's internal features to confirm it. A plausible explanation is that the model associates words like ``roundabout'' and ``baseball diamond'' mainly with side-view imagery, so their textual representation may not align well with the top-down appearance of these objects. The word ``diamond'' may compound this further, since it also carries an unrelated, common meaning. We did not measure feature-space distances, so this remains a hypothesis rather than a confirmed mechanism.

The same two groups apply here. Staying inside the group that only scores present classes, adding text to a box raises aggregate accuracy from 32.81 mAP$_{50:95}$ under Box-Only to 36.07 mAP$_{50:95}$ under Text + Box + Oracle, a gain of 3.26 points (\reftable{tab:cross_modal_effect}), larger than on DIOR and in the opposite direction, with an interval that again excludes zero. The gain is broad rather than driven by a single class. 12 of the 15 classes improve (\reftable{tab:cross_modal_per_class}), and it is largest for football field ($+17.54$) and roundabout ($+9.71$), while only baseball diamond, harbor and ship decrease, each by less than 1.1 points. The plain Text + Box result, 35.35 mAP$_{50:95}$, is a bit lower, since some of that gain is lost to false positives on classes that are not present. Adding text to a box helps on iSAID and hurts on DIOR, so the effect depends on the task rather than one general pattern. Observing the roundabout in \reffig{fig:isaid_roundabout}, a purely visual prompt causes the model to segment the surrounding roads. Similarly, for the harbor in \reffig{fig:isaid_harbor}, the visual prompt alone segments the nearby houses, while the textual prompt alone mistakenly segments the water. When both text and visual prompts are mixed, the model correctly segments the roundabout and perfectly outlines the harbor docks.

\begin{table*}
\caption{Effect of adding text to a box prompt on DIOR and iSAID, measured as the change from Box-Only (B) to Text + Box + Oracle (TBF), which both score only the classes present in the image and share the same exemplar in every run. Values are seven-seed means. $\Delta$ is TBF minus B in points, with its 95\% paired bootstrap interval over test images (1000 resamples, the same resample for both configurations and all seven seeds) in brackets. The class-wise rows refer to AP$_{50:95}$ and give the number of classes that improve, the median class change in points, and the number of classes whose interval excludes zero (positive/negative).}
\label{tab:cross_modal_effect}
\centering
\begin{tabular}{|l|cc|cc|}
\hline
\multirow{2}{*}{\textbf{Metric}} & \multicolumn{2}{c|}{\textbf{DIOR}} & \multicolumn{2}{c|}{\textbf{iSAID}} \\
\cline{2-5}
 & \textbf{mAP$_{50:95}$} & \textbf{mAP$_{50}$} & \textbf{mAP$_{50:95}$} & \textbf{mAP$_{50}$} \\
\hline
B & 67.35 & 80.92 & 32.81 & 57.18 \\
\hline
TBF & 65.90 & 78.94 & 36.07 & 62.43 \\
\hline
$\Delta$ [95\% CI] & $-$1.45 [$-$1.61, $-$1.28] & $-$1.98 [$-$2.22, $-$1.77] & +3.26 [2.78, 3.74] & +5.25 [4.57, 6.10] \\
\hline
Improved & 8/20 & 8/20 & 12/15 & 12/15 \\
\hline
Median & $-$0.58 & $-$0.58 & +1.71 & +1.71 \\
\hline
CI $\neq 0$ (+/$-$) & 11 (3/8) & 11 (3/8) & 13 (11/2) & 13 (11/2) \\
\hline
\end{tabular}
\end{table*}

\begin{table*}
\caption{Class-wise effect of adding text to a box prompt on DIOR (left) and iSAID (right). B and TBF are the seven-seed mean AP$_{50:95}$ (\%) of a single class under Box-Only and Text + Box + Oracle, $\Delta$ is TBF minus B in points, and the last column is the 95\% paired bootstrap interval of $\Delta$ over test images (1000 resamples). $\Delta$ is in bold when the interval excludes zero.}
\label{tab:cross_modal_per_class}
\centering
\scriptsize
\begin{minipage}[t]{0.49\textwidth}
\centering
\begin{tabular}[t]{|l|c|c|c|c|}
\hline
\textbf{Class} & \textbf{B} & \textbf{TBF} & $\boldsymbol{\Delta}$ & \textbf{95\% CI} \\
\hline
Expressway service area & 71.38 & 68.73 & \textbf{$-$2.65} & [$-$3.45, $-$1.72] \\
Expressway toll station & 84.05 & 83.36 & $-$0.69 & [$-$1.37, 0.03] \\
Airplane & 66.49 & 65.99 & \textbf{$-$0.50} & [$-$0.85, $-$0.21] \\
Airport & 92.43 & 92.44 & 0.01 & [$-$0.44, 0.41] \\
Baseball field & 78.32 & 63.12 & \textbf{$-$15.20} & [$-$16.36, $-$13.92] \\
Basketball court & 71.15 & 71.78 & 0.63 & [$-$0.17, 1.44] \\
Bridge & 39.03 & 39.79 & \textbf{0.76} & [0.26, 1.18] \\
Chimney & 74.34 & 73.51 & $-$0.83 & [$-$1.75, 0.04] \\
Dam & 82.86 & 82.34 & $-$0.52 & [$-$1.36, 0.15] \\
Golf field & 87.50 & 86.71 & $-$0.79 & [$-$1.75, 0.20] \\
Ground track field & 79.60 & 76.08 & \textbf{$-$3.51} & [$-$4.16, $-$2.86] \\
Harbor & 49.41 & 54.99 & \textbf{5.58} & [4.40, 6.68] \\
Overpass & 50.83 & 50.99 & 0.15 & [$-$0.46, 0.82] \\
Ship & 50.33 & 45.45 & \textbf{$-$4.88} & [$-$5.73, $-$4.22] \\
Stadium & 92.10 & 92.20 & 0.09 & [$-$0.39, 0.72] \\
Storage tank & 49.19 & 49.75 & \textbf{0.56} & [0.29, 0.90] \\
Tennis court & 78.85 & 78.20 & \textbf{$-$0.65} & [$-$0.95, $-$0.41] \\
Train station & 84.28 & 85.44 & 1.17 & [$-$0.03, 2.49] \\
Vehicle & 33.91 & 28.46 & \textbf{$-$5.45} & [$-$6.33, $-$4.56] \\
Windmill & 30.86 & 28.62 & \textbf{$-$2.24} & [$-$2.54, $-$1.93] \\
\hline
\end{tabular}
\end{minipage}\hfill
\begin{minipage}[t]{0.49\textwidth}
\centering
\begin{tabular}[t]{|l|c|c|c|c|}
\hline
\textbf{Class} & \textbf{B} & \textbf{TBF} & $\boldsymbol{\Delta}$ & \textbf{95\% CI} \\
\hline
Large vehicle & 28.60 & 31.91 & \textbf{3.31} & [2.78, 3.82] \\
Small vehicle & 7.96 & 9.04 & \textbf{1.09} & [0.93, 1.23] \\
Storage tank & 37.46 & 39.18 & \textbf{1.71} & [1.06, 2.44] \\
Ship & 27.85 & 27.14 & \textbf{$-$0.71} & [$-$1.40, $-$0.12] \\
Harbor & 25.35 & 24.50 & \textbf{$-$0.85} & [$-$1.33, $-$0.40] \\
Football field & 46.40 & 63.94 & \textbf{17.54} & [14.47, 20.75] \\
Ground track field & 24.54 & 24.75 & 0.21 & [$-$2.19, 2.62] \\
Baseball diamond & 41.95 & 40.86 & $-$1.09 & [$-$2.67, 0.69] \\
Swimming pool & 28.76 & 30.07 & \textbf{1.31} & [0.55, 2.24] \\
Roundabout & 54.69 & 64.41 & \textbf{9.71} & [6.40, 13.09] \\
Tennis court & 65.67 & 71.35 & \textbf{5.69} & [2.13, 9.56] \\
Basketball court & 46.24 & 49.88 & \textbf{3.65} & [2.09, 5.29] \\
Plane & 31.46 & 33.41 & \textbf{1.95} & [1.42, 2.50] \\
Helicopter & 7.68 & 11.79 & \textbf{4.11} & [2.05, 6.73] \\
Bridge & 17.54 & 18.87 & \textbf{1.34} & [0.74, 2.02] \\
\hline
\end{tabular}
\end{minipage}
\end{table*}

Taken together, these analyses address whether the opposite directions could come from the particular test images or from which classes are present. The intervals in \reftable{tab:cross_modal_effect} exclude zero on both datasets, so neither direction depends on the images that happen to be sampled. The class-wise changes show that the two effects are built differently. On iSAID the gain is spread over most classes and is still $+1.67$ points after the two largest gains are removed. On DIOR the drop is concentrated, since baseball field, vehicle, and ship account for most of it and the remaining 17 classes average only $-0.20$ points. The direction also does not follow class identity, since harbor gains $5.58$ points on DIOR and loses $0.85$ on iSAID, tennis court loses $0.65$ on DIOR and gains $5.69$ on iSAID, and ground track field loses $3.51$ on DIOR and gains $0.21$ on iSAID. We therefore describe the effect of adding text to a box as task-dependent, and not as a general degradation. We did not test whether the difference comes from the task itself, boxes against masks, or from how the two datasets are annotated, so we do not attribute it further.

A plausible explanation is that the visual and textual signals correct different kinds of error. The visual appearance of a harbor may be close to a building's, since they share similar materials and colors, which could mislead a visual-only prompt toward nearby buildings. The word ``harbor,'' on the other hand, may be more closely associated with water, which could mislead a text-only prompt toward the surrounding water instead. Combining the two may let each correct the other's mistake, the visual signal steering the prediction away from water and the text steering it away from buildings, landing on the harbor itself. The same may hold for the roundabout, where the text could help pull the prediction away from the surrounding roads. We did not analyze the model's internal representations to confirm this.

Yet, this mixed prompting can also introduce interference, as observed with the vehicles in \reffig{fig:isaid_harbor}. When queried with only the textual prompts for ``small vehicle'' and ``large vehicle,'' the model is better able to distinguish between the two concepts. However, when queried with visual prompts, or a mix of visual and textual prompts, the model segments both small and large vehicles indiscriminately. 

A plausible explanation is that the words ``small'' and ``large'' are further apart in meaning than the corresponding vehicles are in top-down appearance, which may let the model distinguish the two concepts by text alone but not by sight. When mixing the visual and textual prompts, the text may then be treated as less informative than the shared visual appearance of ``vehicle,'' so the decoder could lean on the visual signal and lose the size distinction. We did not analyze the model's internal representations to confirm this.

\subsubsection{Presence Head Hallucinations and Filtering}
Comparing the unfiltered and oracle-filtered configurations shows how much the decoupled presence head's own errors are costing it, since the oracle configurations use ground-truth class presence rather than the model's own predictions. The unfiltered pipelines show clear false positives, incorrectly extending masks onto background regions. 

Observing the visual outputs, when the model is prompted with the full vocabulary of iSAID concepts, it hallucinates incorrect classes over distinct visual features. In \reffig{fig:isaid_sports}, the model mistakenly segments the tennis courts as basketball courts. Similarly, in \reffig{fig:isaid_baseball}, the model incorrectly detects the generic green grass of the baseball outfield as a ground track field. 

A plausible explanation is a mismatch between the textual and visual sides of these concepts. Rectangular sports courts such as tennis and basketball courts look highly similar from directly overhead, so once ``basketball court'' is introduced as a text prompt, the model may struggle to tell the two apart by their subtle geometric differences. Similarly, ``ground track field'' may be associated too broadly with generic green terrain, leading the model to predict a track where only grass is present. We did not analyze the model's internal representations to confirm this.

Applying the oracle negative filters actively suppresses this background noise. As seen in the oracle configurations of \reffigs{fig:isaid_sports}{fig:isaid_baseball}, the false-positive masks for the basketball courts and ground track fields are completely removed. This filtering lifts the zero-shot Text-Only mode to 12.70 mAP$_{50:95}$ and 24.15 mAP$_{50}$ and the  multimodal setup to 36.07 mAP$_{50:95}$ and 62.43 mAP$_{50}$. However, the small margin of improvement ($\sim1.5$ to $2$ points) suggests that although the presence head introduces false-positive hallucinations, the main bottleneck is missed detections, which we attribute to a domain gap between the text and the aerial imagery.

\begin{table*}
\caption{Class-Wise Instance Segmentation Evaluation on the 15 Classes of the iSAID Dataset. Each value is the mask Average Precision (\%) of a single class, reported as AP$_{50:95}$ (averaged over IoU thresholds 0.50 to 0.95) and AP$_{50}$ (at IoU 0.50). Modality Legend: T = Text-Only (\refConf{config:text}); TB = Text + Box (\refConf{config:textbox}); B = Box-Only (\refConf{config:box}); TF = Text + Oracle (\refConf{config:text_filt}); TBF = Text + Box + Oracle (\refConf{config:textbox_filt}). TB, B, and TBF columns report mean $\pm$ sample standard deviation over seven independently seeded random exemplar draws, while T and TF use no visual prompt and are deterministic, so they are reported from a single run.}
\label{tab:appendix_isaid_merged}
\centering
\resizebox{\textwidth}{!}{
\begin{tabular}{|l|c|c|c|c|c|c|c|c|c|c|}
\hline
& \multicolumn{5}{c|}{\textbf{AP$_{50:95}$}} & \multicolumn{5}{c|}{\textbf{AP$_{50}$}} \\
\cline{2-11}
\textbf{Class Concept} & \textbf{T} & \textbf{TB} & \textbf{B} & \textbf{TBF} & \textbf{TF} & \textbf{T} & \textbf{TB} & \textbf{B} & \textbf{TBF} & \textbf{TF} \\
\hline
Large Vehicle & 26.84 & 31.27 $\pm$ 0.13 & 28.60 $\pm$ 0.35 & 31.91 $\pm$ 0.13 & 28.04 & 47.13 & 57.63 $\pm$ 0.35 & 52.42 $\pm$ 0.70 & 59.11 $\pm$ 0.37 & 49.60 \\
Baseball Diamond & 0 & 40.86 $\pm$ 0.45 & 41.95 $\pm$ 0.70 & 40.86 $\pm$ 0.45 & 0 & 0 & 70.68 $\pm$ 1.07 & 76.51 $\pm$ 0.81 & 70.68 $\pm$ 1.07 & 0 \\
Basketball Court & 0.25 & 44.23 $\pm$ 0.34 & 46.24 $\pm$ 0.51 & 49.88 $\pm$ 0.47 & 1.40 & 0.61 & 64.61 $\pm$ 0.81 & 70.02 $\pm$ 1.13 & 73.87 $\pm$ 0.92 & 3.30 \\
Bridge & 0.52 & 17.53 $\pm$ 0.24 & 17.54 $\pm$ 0.25 & 18.87 $\pm$ 0.26 & 1.40 & 1.16 & 41.96 $\pm$ 0.74 & 42.72 $\pm$ 0.63 & 46.20 $\pm$ 0.86 & 3.49 \\
Football Field & 20.40 & 63.81 $\pm$ 0.50 & 46.40 $\pm$ 0.69 & 63.94 $\pm$ 0.51 & 21.88 & 27.57 & 82.94 $\pm$ 0.85 & 62.82 $\pm$ 0.76 & 83.14 $\pm$ 0.88 & 29.54 \\
Ground Track Field & 0.01 & 23.25 $\pm$ 0.35 & 24.54 $\pm$ 0.30 & 24.75 $\pm$ 0.38 & 0.09 & 0.02 & 39.56 $\pm$ 0.60 & 41.85 $\pm$ 0.59 & 42.17 $\pm$ 0.72 & 0.27 \\
Harbor & 4.00 & 24.45 $\pm$ 0.35 & 25.35 $\pm$ 0.57 & 24.50 $\pm$ 0.36 & 4.18 & 13.15 & 59.61 $\pm$ 0.90 & 59.91 $\pm$ 1.45 & 59.75 $\pm$ 0.91 & 13.64 \\
Helicopter & 2.26 & 11.79 $\pm$ 0.67 & 7.68 $\pm$ 0.29 & 11.79 $\pm$ 0.67 & 2.27 & 13.16 & 44.09 $\pm$ 1.82 & 26.26 $\pm$ 1.03 & 44.09 $\pm$ 1.82 & 13.19 \\
Plane & 29.62 & 33.36 $\pm$ 0.14 & 31.46 $\pm$ 0.23 & 33.41 $\pm$ 0.14 & 29.69 & 80.98 & 83.66 $\pm$ 0.32 & 78.77 $\pm$ 0.78 & 83.83 $\pm$ 0.33 & 81.21 \\
Roundabout & 8.14 & 64.13 $\pm$ 0.56 & 54.69 $\pm$ 0.49 & 64.41 $\pm$ 0.60 & 11.72 & 12.84 & 85.81 $\pm$ 0.58 & 73.69 $\pm$ 0.55 & 86.39 $\pm$ 0.63 & 18.42 \\
Ship & 8.35 & 27.06 $\pm$ 0.61 & 27.85 $\pm$ 0.59 & 27.14 $\pm$ 0.61 & 8.50 & 14.05 & 51.63 $\pm$ 1.24 & 52.71 $\pm$ 0.92 & 51.80 $\pm$ 1.24 & 14.30 \\
Small Vehicle & 6.76 & 9.00 $\pm$ 0.03 & 7.96 $\pm$ 0.05 & 9.04 $\pm$ 0.03 & 6.96 & 18.33 & 23.80 $\pm$ 0.04 & 20.45 $\pm$ 0.07 & 23.92 $\pm$ 0.04 & 18.60 \\
Storage Tank & 19.67 & 38.96 $\pm$ 0.25 & 37.46 $\pm$ 0.41 & 39.18 $\pm$ 0.24 & 20.42 & 30.73 & 63.79 $\pm$ 0.74 & 60.74 $\pm$ 0.90 & 64.25 $\pm$ 0.73 & 31.94 \\
Swimming Pool & 18.28 & 30.05 $\pm$ 0.31 & 28.76 $\pm$ 0.42 & 30.07 $\pm$ 0.31 & 18.31 & 42.37 & 62.46 $\pm$ 0.27 & 59.26 $\pm$ 0.63 & 62.51 $\pm$ 0.27 & 42.45 \\
Tennis Court & 32.70 & 70.45 $\pm$ 0.53 & 65.67 $\pm$ 1.45 & 71.35 $\pm$ 0.55 & 35.59 & 38.86 & 83.47 $\pm$ 0.74 & 79.50 $\pm$ 1.69 & 84.69 $\pm$ 0.78 & 42.33 \\
\hline
\end{tabular}
}
\end{table*}

\subsubsection{Granular Class Analysis: Persistent Vulnerabilities and Structural Drops}

A rigorous, class-by-class examination of the exhaustive data grid in \reftable{tab:appendix_isaid_merged} strips away the statistical smoothing of macro averages, exposing severe, localized failures that persist across the SAM~3 model:

\begin{itemize}
    \item \textbf{Complete Semantic Deficits:} Observing the data for complex categories like \textit{Baseball Diamond}, performance collapses to an absolute 0 AP$_{50:95}$ under the pure zero-shot text Configuration (\refConf{config:text}). As visually confirmed in \reffig{fig:isaid_baseball}, the model simply returns empty predictions. A plausible explanation is that the textual concept of a baseball diamond and its top-down visual appearance are too far apart for the model to connect from text alone. Adding a visual prompt (\refConf{config:textbox_filt}) anchors the target directly and raises the score to 70.68 AP$_{50}$, and the contrast between the two results highlights what appears to be an acute semantic blind spot for this class. We did not analyze the model's internal representations to confirm this interpretation.

    \item \textbf{Micro-Scale and Clutter Vulnerabilities:} A critical structural weakness is observed in the \textit{Small Vehicle} category, which scores a severely depressed 6.76 AP$_{50:95}$ in zero-shot text mode and stalls heavily at 9.04 AP$_{50:95}$ under optimal multimodal filtering. Observing \reffig{fig:isaid_harbor}, the model constantly conflates small and large vehicles regardless of the prompt. A plausible explanation is a resolution limitation. Densely packed, sub-pixel targets may lose distinctive visual details during the Image Encoder's pooling stages, which could make small and large vehicles visually indistinguishable to the model at this scale. We did not analyze the model's internal features to confirm this.

    \item \textbf{Amorphous Boundary Failures:} Visually complex, poorly bounded regions like \textit{Harbor} display an acute resistance to high-precision segmentation, scoring a low 4.00 AP$_{50:95}$ in text mode and managing only 24.50 AP$_{50:95}$ under optimal multimodal conditions. As observed in \reffig{fig:isaid_harbor}, the visual boundaries of a harbor blend into adjacent water and building structures. A plausible explanation is that the model's visual representation of a harbor overlaps with these surrounding elements, which would be consistent with the boundary confusion observed at the pixel level. While mixing modalities successfully shifts the core prediction toward the docks, this confusion persists to some degree under every configuration we tested. We did not analyze the model's internal features to confirm this.
\end{itemize}

\begin{table}[h]
\caption{Size-stratified segmentation AP for selected iSAID classes, comparing the weakest configuration (Text-Only) against the strongest (Text + Box + Oracle). AP$_S$, AP$_M$, AP$_L$ denote AP for small, medium, and large objects by pixel area, following the COCO definition. Large Vehicle and Small Vehicle contain no large-sized ground-truth instances, shown as --. The All classes row reports the mean across classes, mAP$_S$, mAP$_M$, and mAP$_L$. Text + Box + Oracle values are reported as mean $\pm$ sample standard deviation over seven independently seeded random exemplar draws.}
\label{tab:isaid_size_stratified}
\centering
\resizebox{\columnwidth}{!}{
\begin{tabular}{|l|c|c|c|c|c|c|}
\hline
\multirow{2}{*}{\textbf{Class}} & \multicolumn{3}{c|}{\textbf{Text-Only}} & \multicolumn{3}{c|}{\textbf{Text + Box + Oracle}} \\
\cline{2-7}
 & AP$_S$ & AP$_M$ & AP$_L$ & AP$_S$ & AP$_M$ & AP$_L$ \\
\hline
All classes & 6.29 & 18.20 & 23.00 & 25.86 $\pm$ 0.31 & 49.87 $\pm$ 0.14 & 52.76 $\pm$ 0.52 \\
Baseball Diamond & 0.00 & 0.00 & 0.00 & 28.14 $\pm$ 1.92 & 53.28 $\pm$ 0.72 & 37.45 $\pm$ 0.67 \\
Small Vehicle & 7.21 & 2.89 & -- & 9.24 $\pm$ 0.06 & 21.63 $\pm$ 1.54 & -- \\
Large Vehicle & 15.18 & 54.04 & -- & 19.61 $\pm$ 0.16 & 58.26 $\pm$ 0.38 & -- \\
Harbor & 2.75 & 9.84 & 3.69 & 23.46 $\pm$ 1.15 & 35.17 $\pm$ 0.34 & 16.54 $\pm$ 0.33 \\
\hline
\end{tabular}
}
\end{table}

Table~\ref{tab:isaid_size_stratified} shows the same pattern for iSAID. Small Vehicle's AP$_S$ (9.24 under Text + Box + Oracle) is close to its class-wise AP$_{50:95}$ in \reftable{tab:appendix_isaid_merged}, since nearly all of its ground-truth instances are small, while its AP$_M$ reaches 21.63, more than double. Large Vehicle shows a comparable gap between AP$_S$ and AP$_M$ (19.61 versus 58.26), indicating this is a general effect of object size rather than one specific to how a class is named. As with DIOR, we did not test multiple input resolutions directly, so this remains supporting evidence for the resolution hypothesis rather than confirmation of it.

\subsubsection{Comparison Against General-Domain Open-Vocabulary Segmentation Models}
\reftable{tab:ovd_baseline_comparison} also reports segmentation performance on iSAID for the same set of general-domain, open-vocabulary models, evaluated under the protocol described in Section~\ref{subsec:eval_protocol}. SAM~3 again leads every baseline, beating the strongest one, OWLv2's ensemble variant, by 6.59 points at mAP$_{50}$ and 4.65 points at mAP$_{50:95}$, a proportionally larger gap than SAM~3 achieves on DIOR's detection task. Instance segmentation adds a mask-quality dimension that plain detection does not, and the size of this gap suggests SAM~3's advantage is not limited to finding objects but extends to delineating them.

For YOLOE, the only baseline model here capable of producing masks natively, converting its own predicted boxes through SAM~1 outperforms its purpose-trained segmentation head at both checkpoint sizes, 3.03 versus 2.84 mAP$_{50}$ at the large size and a similar margin at the small size. This is a useful sanity check in its own right. Even a general-purpose segmenter given a reasonably placed box currently produces sharper masks on this domain than a lightweight model's own dedicated mask branch, reinforcing that box quality, not mask decoding, is the harder half of this problem.

Ground-truth presence filtering improves SAM~3's text-only result only modestly here too, from 22.73 to 24.15 mAP$_{50}$, echoing the same pattern observed on DIOR (\reftable{tab:isaid_macro_ablation}). SAM~3 is the strongest model at this task among those tested with no domain training and no visual guidance, but its own box-guided results in \reftable{tab:isaid_macro_ablation} remain far higher, leaving a wide margin between what a text prompt can currently achieve on its own and what the model can do once shown what it is looking for.

\section{Conclusion and Future Work}
\label{section:Conclusion and Future Work}
This study presents one of the first comprehensive, multi-task evaluations of the Segment Anything Model 3 (SAM~3) in the Earth Observation domain. We benchmarked the model across scene classification (AID), open-vocabulary object detection (DIOR), and instance segmentation (iSAID) without any fine-tuning. Our findings show that SAM~3 has promising potential for remote sensing. Its visual prompt reliably localizes objects once shown what to look for, and its text-only performance leads every general-domain, open-vocabulary model we compared it against on DIOR and iSAID. At the same time, its absolute accuracy from text alone remains limited, pointing to a persistent gap between remote sensing vocabulary and the model's general-domain visual representations.

Our five-configuration ablation across prompt types points to where this gap shows up. When given a visual prompt, an image exemplar that reveals one true instance of the target, SAM~3 reliably localizes and outlines the remaining objects. Text alone performs far worse, and combining text with a visual prompt raises accuracy on iSAID but lowers it on DIOR, so the effect is task-dependent rather than a single general pattern. A plausible explanation is that the model's text-to-concept associations, learned mostly from ground-level imagery, do not transfer cleanly to the top-down appearance of aerial objects, though we did not analyze the model's internal features to confirm this. To quantify how sensitive these visual-prompt results are to which ground-truth instance is drawn as the exemplar, we repeated every visual-prompt configuration seven times with independently seeded random draws on both datasets and report mean $\pm$ sample standard deviation throughout. The mAP$_{50:95}$ for these configurations is stable within about 0.2 points on both datasets, and the large majority of per-class and size-stratified cells are stable within about 1 point, with a small number of exceptions where a class's ground truth is sparse or visually heterogeneous, most notably DIOR Vehicle's AP$_L$ ($49.63 \pm 5.17$). The text prompts are bare class names, which matches the short noun phrases SAM~3 is trained on, so we did not vary prompt phrasing. The failure patterns discussed throughout this study were identified from selected qualitative examples and the class-wise tables. We did not quantify how often each pattern occurs across the full test set, which we leave for future work.

These findings suggest a direction for future work. Since SAM~3's visual prompt already localizes aerial objects well, the largest gains are likely to come from lightweight, parameter-efficient fine-tuning, such as Low-Rank Adaptation (LoRA) or vision-language contrastive realignment, targeting the text encoder and multimodal decoder rather than retraining the full visual backbone. Aligning remote sensing vocabulary with top-down visual representations in this way could narrow the gap between SAM~3's text-only and visual-prompt performance, without the cost of full fine-tuning.

\section*{Acknowledgment}
Google Gemini and Claude were used solely for the purpose of English language editing throughout the manuscript.

\bibliographystyle{IEEEtran}
\bibliography{references}

\end{document}